\documentclass[1column]{article}
\usepackage[pdftex]{graphicx}
\usepackage[a4paper, total={6in, 8in}]{geometry}
\usepackage[font=footnotesize, margin=1cm]{caption}
\usepackage{float}
\usepackage{subfig}
\usepackage{epstopdf}
\usepackage{color}
\graphicspath{{bmp/}{png/}{bmp2/}{eps/}}

\usepackage{amssymb}
\usepackage{amsmath}
\usepackage{multirow}

\begin{document}
\title{Geometry-Aware Graph Transforms for Light Field Compact Representation}
\author{Mira~Rizkallah$^{\ast}$, Xin~Su$^\dag$, Thomas~Maugey$^{\dag}$, and~Christine~Guillemot$^\dag$,\\
        $^\ast$ IRISA, Campus Universitaire de Beaulieu, 35042 Rennes, France \\
        $^\dag$ INRIA Rennes Bretagne Atlantique, Rennes, France
\thanks{This work has been supported in part by the EU H2020 Research and Innovation Programme under grant agreement No 694122 (ERC advanced grant CLIM).}}


\maketitle

\begin{abstract}
The paper addresses the problem of energy compaction of dense 4D light fields by designing geometry-aware local graph-based transforms. Local graphs are constructed on super-rays that can be seen as a grouping of spatially and geometry-dependent angularly correlated pixels. Both non separable and separable transforms are considered. 
Despite the local support of limited size defined by the super-rays, the Laplacian matrix of the non separable graph remains of high dimension and its diagonalization to compute the transform eigen vectors remains computationally expensive. 
To solve this problem, we then perform the local spatio-angular transform in a separable manner. We show that when the shape of corresponding super-pixels in the different views is not isometric, the basis functions of the spatial transforms are not coherent, resulting in decreased correlation between spatial transform coefficients. We hence propose a novel transform optimization method that aims at preserving angular correlation even when the shapes of the super-pixels are not isometric. 
Experimental results show the benefit of the approach in terms of energy compaction. A coding scheme is also described to assess the rate-distortion perfomances of the proposed transforms and is compared to state of the art encoders namely HEVC and JPEG Pleno VM 1.1.

\end{abstract}



\section{Introduction}
Recently, there has been a growing interest in light field imaging. By sampling the radiance of light rays emitted by the scene along several directions, light fields enable a variety of post-capture processing techniques such as refocusing, changing perspectives and viewpoints, depth estimation, simulating captures with different depth of fields and 3D reconstruction \cite{WIL05,RNG06,GEO10b}. 
This however comes at the expense of collecting large volumes of redundant high-dimensional data, which appears to be one key downside
of light fields. 

Research effort has been recently dedicated to the design of light field compression algorithms, by either adapting standardized solutions (in particular HEVC) to light
field data (\textit{e.g.}  \cite{liu2016pseudo} \cite{conti2016hevcvideo} \cite{Li2016compression}), by proposing homography-based low rank models for reducing the angular dimension \cite{jiang2017light}, or by investigating local Gaussian mixture models in the 4D ray space 
\cite{Verhack2017}. The authors in \cite{tabuslossy}, use a depth‐based segmentation of the light field into 4D spatio‐angular blocks with prediction followed by JPEG‐2000.

In this paper, we address the problem of graph transforms optimization for light fields energy compaction and compact representation. Indeed, light fields record illumination of light rays emitted by a scene in different orientations. The captured data for a static light field is represented by a 4D function $LF(u, v, x, y)$, and contains redundant information in both the spatial and angular dimensions.
Those correlations could in principle be represented by a {\em huge} non separable graph connecting pixels within and across views of the entire light field. The basis functions of a graph Fourier transform \cite{shuman2013emerging} could then be used 
to decorrelate the color signal residing on the graph vertices. However, such a graph would have a very high number of vertices,
each vertex corresponding to a light ray. This makes the diagonalization of the laplacian matrix unfeasible, hence, the computation of the graph Fourier transform not practical. 

To lower the dimensionality of the problem, we propose to partition the big graph structure into smaller ones that are coherent and correlated inside and across the views. This can be viewed as cutting unreliable edges from the \textit{global} graph. To perform this partitioning, we group similar pixels within and across views based on the concept of super-rays defining the supports of the set of local graph transforms.  The concept of super-ray has been introduced in \cite{Hog2017} as an extension to light fields of the concept of super-pixels.

The authors in \cite{su2018rate} used super-rays as the supports of separable shape-adaptive Discrete Cosine Transform (DCT).  Super-pixels are used in \cite{rizkallah2018graph} as the supports of local graph transforms, and tested in a predictive scheme based on view synthesis. The angular transform is however applied on super-pixels that are co-located on all views, hence
not exploiting scene geometry, due to the difficulty to design separable graph transforms that at the same time follow the scene geometrical information and preserve angular correlations. We come back on this point in the sequel.

In this paper, we address the problem of designing local super-ray based non separable and separable graph transforms following the scene geometry. Towards this goal, we first propose a specific super-ray construction method to limit shape variations of the super-pixels forming a given super-ray. Despite the local support of limited size defined by the super-rays, the Laplacian matrix remains of high dimension and its diagonalization to compute the transform eigen vectors is computationally expensive. 
An intuitive way to solve this problem is to perform the transform in a separable manner: a first spatial transform applied per super-pixel inside each view, then an angular transform between corresponding super-pixels across the views to capture angular dependencies. We have however observed that if the shape of the super-ray undergoes a slight change between views, the basis functions computed from the graph laplacian have very different forms from one super-pixel to the corresponding ones in the other views, resulting in a decreased correlation between spatial transform coefficients. 

The difficulty is therefore how to optimize the spatial transforms applied on each super-pixel of the super-ray in such a way that the angular correlation is well preserved. Preserving angular correlation is important in order to best compact the light field energy. The angular correlation is preserved, only if the eigen vectors of the spatial transforms computed independently on different shapes (the super-pixels forming the super-ray) are reasonably consistent, i.e. only when the shapes of the transform supports are approximately isometric.
We propose in this paper a novel method to optimize the spatial transforms in such a way that the basis functions approximately diagonalize their respective Laplacians while being coherent across the views, given the scene geometry.

Experimental results show that 
the proposed super-ray construction method yields, for the light fields considered in the tests, up to $60$ percent coherent supports out of all super-rays, which facilitates the application of a separable graph transform. The results also show that the optimized separable graph transform yields higher energy compaction, and significant rate-distortion performance gains, compared to the non optimized separable transform, when some super-rays are shape-varying across the views. The proposed simple coding scheme based on these local separable transforms is shown to outperform light field coding schemes based on HEVC and JPEG Pleno at high bitrate following the common test conditions.

In summary our contributions are as follows:
\begin{itemize}
\item We propose local graph transforms based on the concept of super-rays adapted to scene geometry. To define the supports of the local transforms, we first propose (section \ref{sec:SRconstruction}) a new algorithm to segment the light field into super-rays. The method takes as input only the top-left color image and a sparse set of disparities. The resulting segmentation defines the supports of local graph transforms.
\item We then introduce (section \ref{sec:GT}) a novel method to optimize the spatial transforms in such a way that the basis functions are coherent across the views, given the scene geometry. 
\item We analyze the properties in terms of energy compaction of the proposed super-rays based graph transforms. 
\item A complete coding scheme (section \ref{sec:schemes}) is also described to assess the
rate-distortion performances of these novel transforms on a set of real light fields. 
\end{itemize}

\section{Related work}
We first briefly review prior work on graph transforms design for signal (and in particular image) energy compaction, problem related to the core of the paper. For sake of completeness, the proposed transforms being validated in a complete coding scheme, we also give a brief overview of recent work on light field compression.
\subsection{Graph Transforms} 
Recently, graph signal processing has been applied to different image and video coding applications, especially for piecewise smooth images. In \cite{shen2010edge,kim2012graph}, the authors propose a graph-based coding method where the graph weights are defined considering pairwise similarities between pixel intensities. Another efficient graph construction method has been proposed in \cite{hu2015multiresolution} for piecewise smooth images. 
For each signal in a block, they select the Graph Fourier Transform minimizing the rate distortion cost. A signed graph Fourier transform has also been proposed in \cite{su2017graphfourier} for depth map coding, accounting for negative weights between pixels.

For natural images, most of the work has focused on designing sparse graphs or using graph templates that capture principal gradient-based structures in images \cite{pavez2015gtt}\cite{rotondo2015designing}. This is mostly useful in textured images.  While most of the aforementioned transform coding strategies did not account for the graph coding cost, in a later work \cite{fracastoro2016graph}, a rate-distortion optimized graph learning approach has been proposed to code natural images while taking into account both the sparsity of the transformed coefficients and the graph coding cost.
Several graph based approaches have also been proposed to code intra and inter predicted residual blocks in video compression, using generalized graph Fourier transform \cite{hu2015intra}, simplified graph templates transforms \cite{egilmez2015graph}, or separate line graph based transforms \cite{lu2016symmetric}.

In this paper, we build graphs that follow the scene geometry and we then propose separable graph based transforms that best exploit light fields spatial and angular correlation.

\subsection{Light Fields Compression} 
Existing light fields compression solutions can be broadly classified into two categories: approaches directly compressing the lenslet images or approaches coding
the views extracted from the raw data.
Methods proposed for compressing the lenslet images mostly extend HEVC intra
coding modes by adding new prediction modes to exploit similarity between lenslet
images (\textit{e.g.}  \cite{conti2012new}, \cite{conti2016hevc}, \cite{conti2016hevcvideo},  \cite{Li2016compression}). The authors in \cite{tabuslossy} propose a lenslet-based compression scheme that uses depth, disparity and sparse prediction followed by JPEG-2000 residue coding.

A second category of methods consists in encoding the set of views which can be extracted from the lenslet images after de-vignetting, demosaicing and alignment of the micro-lens array on the sensor, following e.g. the raw data decoding pipeline in \cite{David2017white}. Several methods code the views as pseudo video sequences using HEVC \cite{liu2016pseudo}, \cite{rizkallah2016impact}, or the latest JEM coder \cite{Jia2017}, or extend HEVC to multi-view coding \cite{Ahmad2017}. Low rank models as well as local Gaussian mixture models in the 4D rays space are proposed in \cite{jiang2017light} and \cite{verhack2017steered} respectively. View synthesis based predictive coding has also been investigated in \cite{zhao2017light} where the authors use a linear approximation computed with Matching Pursuit
for disparity based view prediction. The authors in \cite{jiang2017hot3D} and \cite{su2017graph} use instead a the convolutional neural network (CNN) architecture proposed in \cite{kalantari2016learning} for view synthesis and prediction. The prediction residue is then coded using HEVC \cite{jiang2017hot3D}, or using local residue transforms (SA-DCT) and coding \cite{su2017graph}. 
The proposed transforms could also be used for residue coding. However, to best assess their de-correlation advantage, in the experiments reported below, they are directly applied on the color values of the entire 4D light field data.

\section{Super-rays and Graph construction}
\label{sec:SRconstruction}
The compression efficiency of any coder based on block partitioning and transform coding does undeniably depend on the way the partitioning is done, and on how the resulting segmentation adheres to object boundaries. While traditional transforms such as 2D DCT applied on a square or rectangular support may fail due to high frequencies captured on the object boundaries, here we rely on a segmentation of the entire 4D light field into super-rays.

\subsection{Light field Segmentation in Super-Rays}
\begin{figure}[h]
  \centering
  \includegraphics[width=0.45\textwidth]{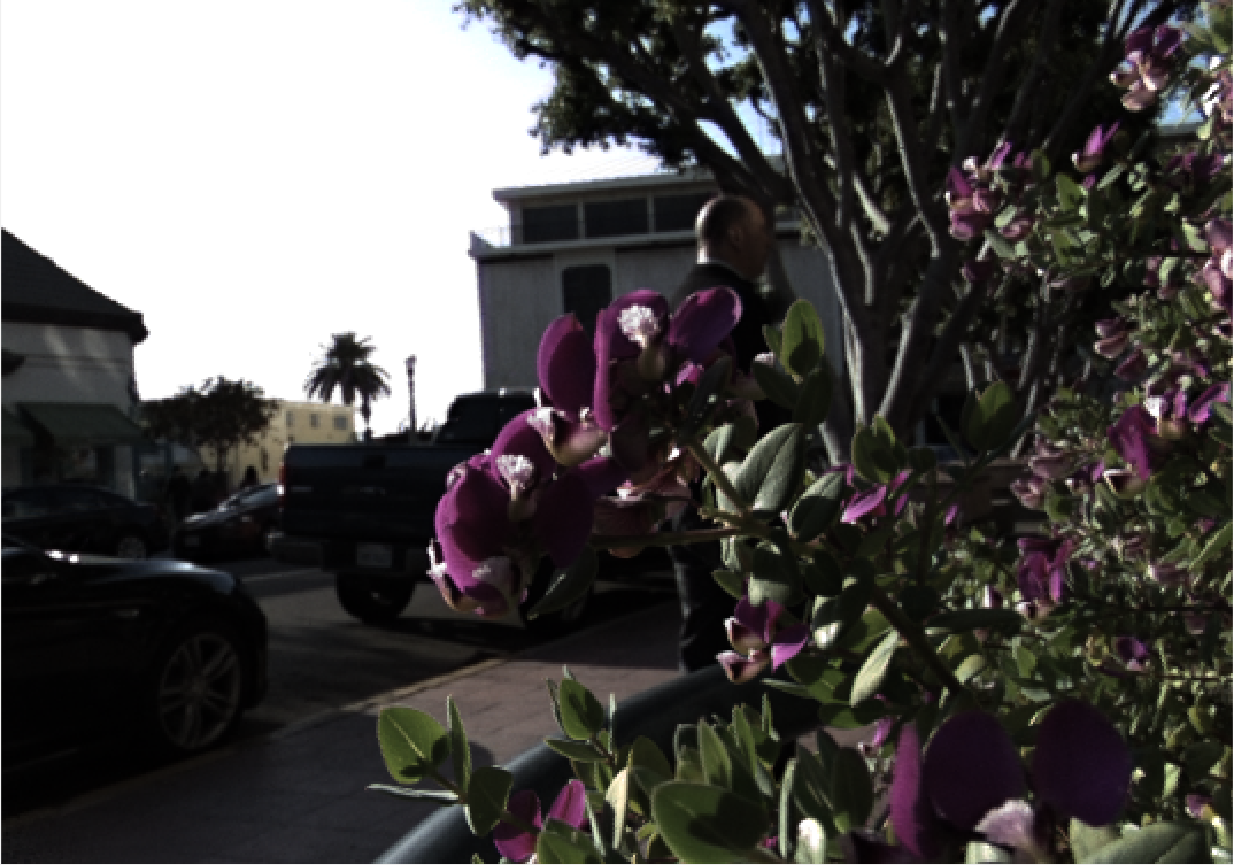}
  \includegraphics[width=0.45\textwidth]{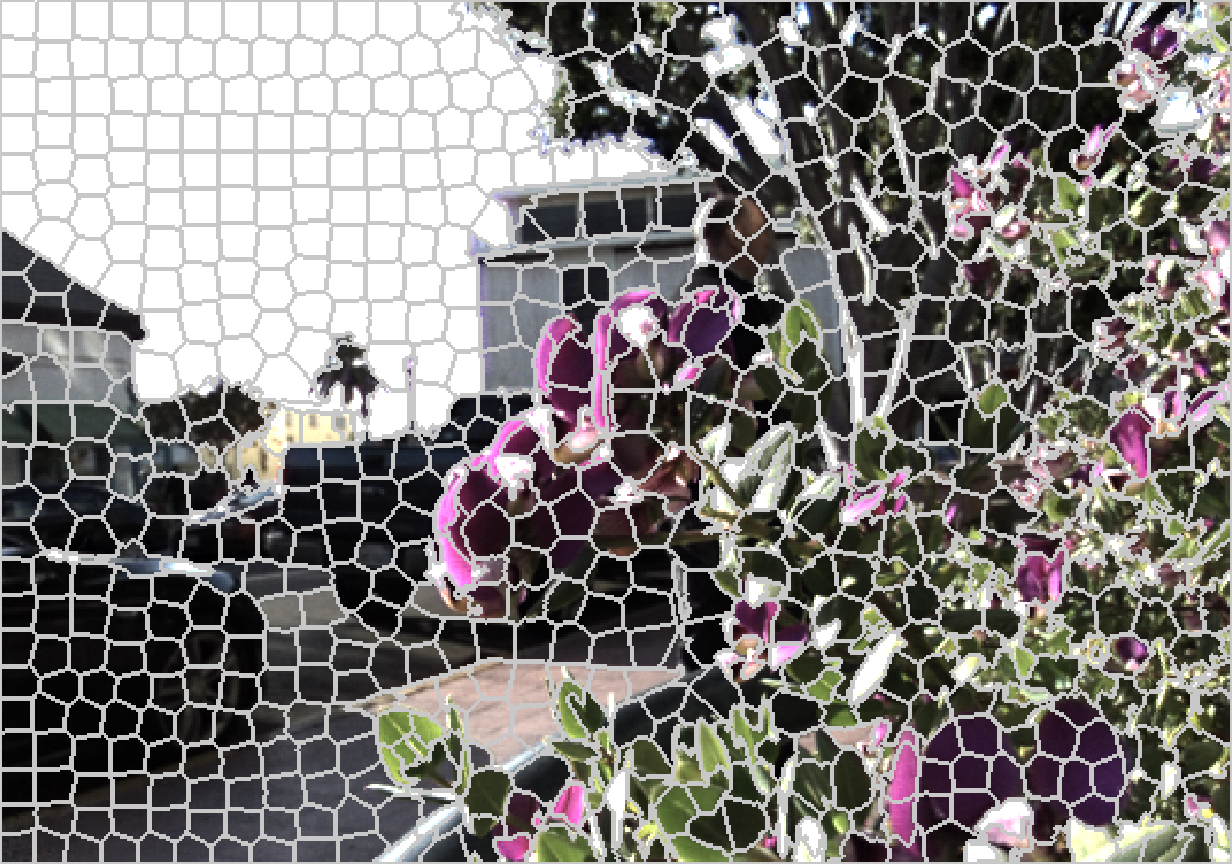}
\caption{The original view $I_{1,1}$ of \emph{Flower1} (left) and the corresponding super-pixel segmentation (right).
}
\label{fig:SPExample}
\end{figure}

\begin{figure}[h]
  \centering
  \includegraphics[width=0.9\textwidth]{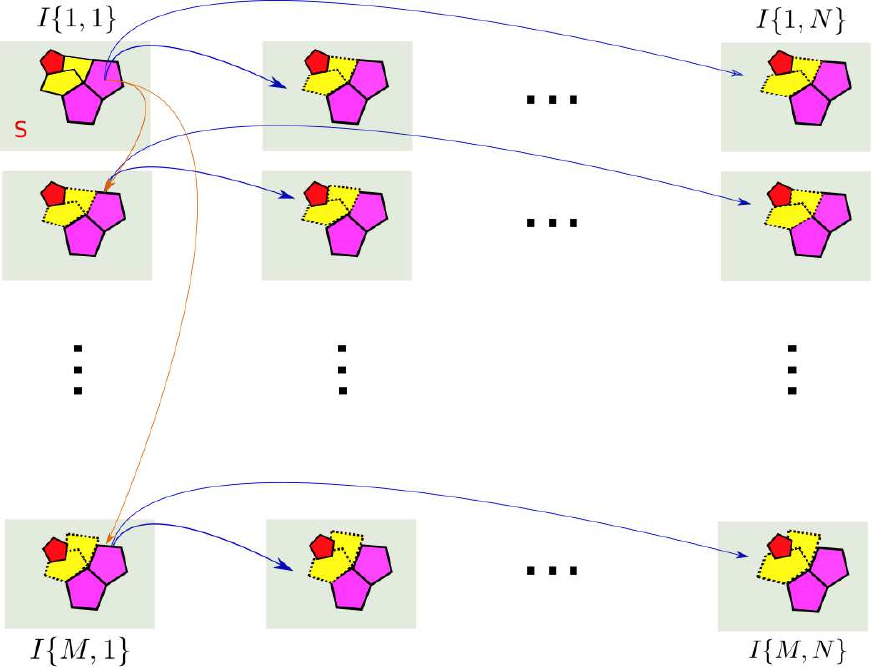}
\caption{Image showing the super-ray construction. The algorithm proceeds row by row. In the first row, only horizontal projections (blue arrows) are performed. In every other row, first a vertical projection (red arrow) than $N-1$ horizontal projections (blue arrows) are performed.}
\label{fig:srconstruction}
\end{figure}

The concept of super-ray has been introduced in \cite{hog2017superrays} as an extension of super-pixels \cite{achanta2012slic} to group light rays coming from the same 3D object, \textit{i.e.} to group pixels having similar color values and being close spatially in the 3D space. The method performs a k-means clustering of all light rays based on color and distance in the 3D space.
To deal with dis-occlusions, a slightly modified formulation is proposed in \cite{su2018rate} where the dense depth information is also used in the clustering. When the depth information is not fully reliable, this method results in inconsistent super-rays across views.
In addition, the signalling cost of such a global light field segmentation is high.
In order to make the super-rays more consistent across the views, we suggest a modified version where we compute super-pixels in the top-left view as shown in Figure \ref{fig:SPExample}. Then, using the disparity map, we project the segmentation labels to all the other views. Namely, having a segmentation map in the top left view and the corresponding disparity map, we compute the median disparity per super-pixel, and use it to project the segmentation mask to the other views. More precisely, the algorithm proceeds row by row. In the first row of views, we perform horizontal projections from the top-left $I\{1,1\}$ to the $N-1$ views next to it. For each other row of views, a vertical projection is first carried out from the top view $I\{1,1\}$ to recover the segmentation on view $I\{m,1\}$, then $N-1$ horizontal projections from $I\{m,1\}$ to the $N-1$ other views are performed, as shown in Figure \ref{fig:srconstruction}. 

An example of segmentation $S$ is shown in Figure \ref{fig:srconstruction}, where the background consists of two yellow superpixels, and two foreground objects are labeled with red and pink. The disparity of the two objects is equal to $1$, while the background is almost fixed with a disparity equal to $0$. At the end of each projection, some shapes are projected in all the views without interfering with others. Those typically represent flat regions inside objects (for example, the object labeled in pink). While others, mainly consisting of occluded and occluding segments end up superposed in some views, for example, the red object occluding pixels from the yellow background. In this case, the occluded pixels are assigned the label (e.g. red) of the neighboring super-ray corresponding to the foreground objects (\textit{i.e.} having the higher disparity). As for appearing pixels, for example, between the yellow background and pink object, they will be clustered with the background super-rays (\textit{i.e.} having the lower disparity e.g. yellow). The super-rays that end up with different shapes in the views are marked with a dashed contour.

\subsection{Graph Construction}
\begin{figure}[h]
  \centering
  \includegraphics[width=0.9\textwidth]{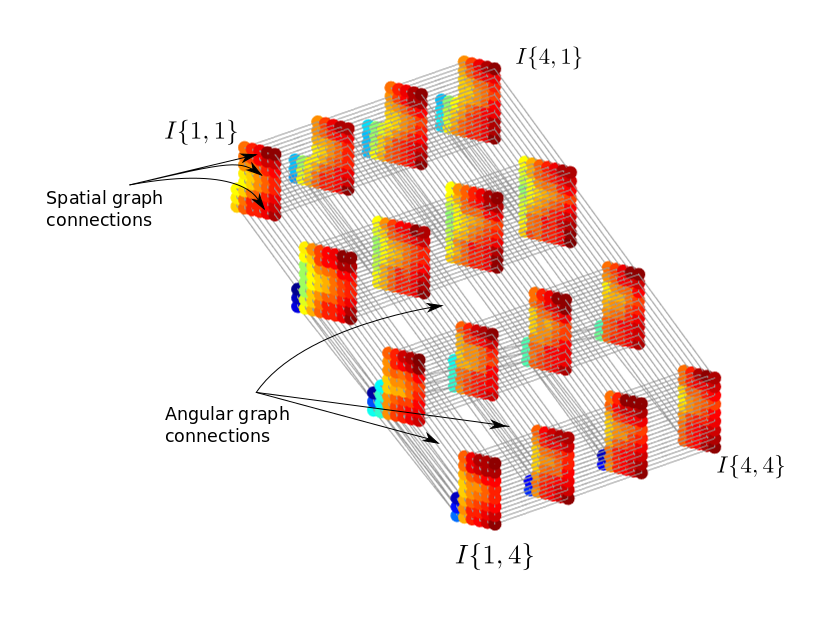}
\caption{Example of local non-separable graph within a super-ray. We can see the connections within super-pixels in each view, as well as connections between pixels belonging to different views. The color assigned to the vertices is the luminance value of the pixels. For visualization purposes, we show the luminance in false colors.}
\vspace{-0.5cm}
\label{fig:SPExample_COLOR}
\end{figure}

In order to jointly capture spatial and angular correlations between pixels in the light field, we first consider a local non separable graph per super-ray. More precisely, if we consider the luminance values in the whole light field and a segmentation map $S$, the $k^{th}$ super-ray $SR_k$ can be represented by a signal $f_k \in \mathbb{R}^{N}$ defined on an undirected connected graph $\mathcal{G} = \{\mathcal{V},\mathcal{E}\}$ which consists of a finite set $\mathcal{V}$ of vertices corresponding to the pixels at positions $ \{u_l,v_l,x_l,y_l\}, l = 1\dots N$ such that $S(u_l,v_l,x_l,y_l) = k$. A set $\mathcal{E}$ of edges connect each pixel and its 4-nearest neighbors in the spatial domain (i.e. in each view), and to its corresponding pixels, found by disparity based projection, in the 4 nearest neighboring views. An example of graph built inside a super-ray is shown in Figure \ref{fig:SPExample_COLOR}.

\section{Graph Transforms}
\label{sec:GT}
In this section, we focus on the design of suitable transforms for the signals (color or residues) residing on the local graphs defined above. 

\subsection{Non Separable Graph Transform} 
Let us consider the $k^{th}$ super-ray $SR_k$ and its corresponding local graph $\mathcal{G}$.  We start by defining its adjacency matrix $\mathbf{A}$ with 
entries $A_{mn}=1$, if there is an edge $e = (m,n)$ between two vertices $m$ and $n$, and  $A_{mn}=0$ otherwise.
The adjacency matrix is used to compute the Laplacian matrix $\mathbf{L = D - A}$, where $\mathbf{D}$ is a diagonal degree matrix whose $i^{th}$ diagonal element $D_{ii}$ is equal to the sum of the weights of all edges incident to node $i$. The resulting Laplacian matrix $\mathbf{L}$ is symmetric positive semi-definitive and therefore can be diagonalized as: 
\begin{equation}
\mathbf{L = U^\top \Lambda U }
\end{equation}
where $\mathbf{U}$ is the matrix whose rows are the eigenvectors of the graph Laplacian and $\mathbf{\Lambda}$ is the diagonal matrix whose diagonal elements are the corresponding eigenvalues.
The laplacian eigenbases $\mathbf{U}$ are analogous to the Fourier bases in the Euclidean domain and allow representing the signals residing on the graph as a linear combination of eigenfunctions akin to Fourier Analysis. 
This is known as the Graph Fourier transform. For the signal $f_k$ defined on the vertices of the local graph,
the transformed coefficients vector $\hat{f_k}$ is defined in \cite{shuman2013emerging} as:
\begin{equation}
\hat{f_k} = \mathbf{U}f_k
\end{equation}
The inverse graph Fourier transform is then given by
\begin{equation}
f_k= \mathbf{U}^\top \hat{f_k} 
\end{equation}

Although this would be the ideal decorrelating transform for the signal, the Laplacian of such graph, despite the locality, remains of high dimension (almost $6000$ nodes per super-ray) leading to a high transform computational cost. To limit the computational cost, we then consider separable local transforms. 

\begin{figure*}[htp]
  \centering
  \includegraphics[width=0.9\textwidth]{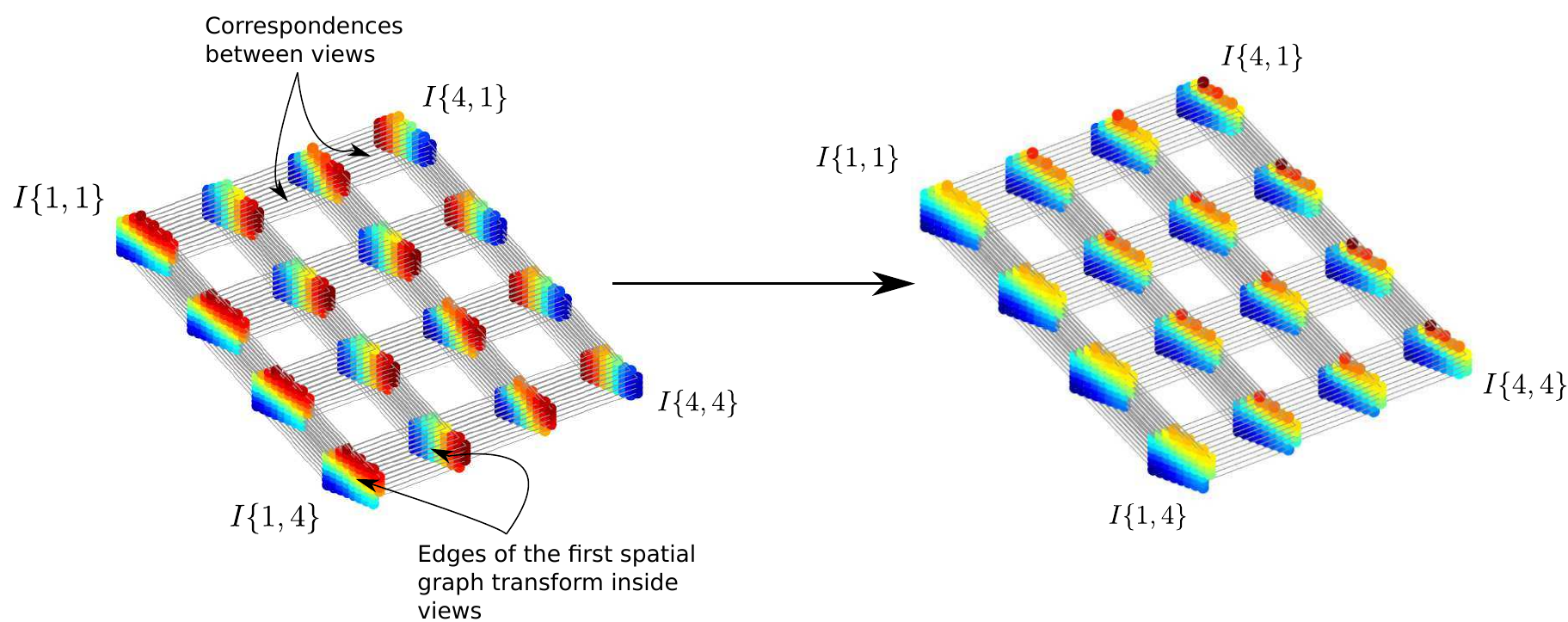}
  \caption{Second eigenvector of different super-pixels belonging to the same super-ray before and after optimization.}
  \vspace{-0.5cm}
  \label{fig:not__opt}
\end{figure*}

\subsection{Coherent Separable Graph Transform} 
\label{sec:CSGT}
The separable graph transform is defined by a first spatial transform followed by a second angular transform as detailed below.

\subsubsection{First spatial graph transform}
If we consider the luminance values in only one sub-aperture image $v$ of the light field and a segmentation map $S$, the $k^{th}$ super-ray  $SR^v_k$ can be represented by a signal $f^v_k \in \mathbb{R}^{N^v_k}$ defined on an local spatial graph with only connections in the spatial domain (\textit{i.e.} between the neighboring pixels in a super-pixel, but not across the views in a super-ray). The matrix ${\mathbf{U}_s}$, being the eigen-vectors of the spatial laplacian $\mathbf{L}_s$, is used to compute the first spatial graph transform :
For the signal $f^v_k$ defined on the vertices of the graph, the transformed coefficients vector $\hat{f^v_k}$ is defined in \cite{shuman2013emerging} as:
\begin{equation}
\hat{f^v_k} = \mathbf{U}_s^\top f^v_k
\end{equation}
The inverse spatial graph Fourier transform is then given by
\begin{equation}
f^v_k= \mathbf{U}_s \hat{f^v_k} 
\end{equation}

\subsubsection{Second angular graph transform}
In order to capture inter-view dependencies and compact the energy into fewer coefficients, we perform a second graph based transform, in the angular dimension. Note that, for a given super-ray, we do not necessarily have the same number of pixels, hence coefficients resulting from the spatial transforms, in all the views. For a given band $b$ (coefficients corresponding to the $b^{th}$ eigenvectors of the spatial transforms), we construct a graph of $N_b$ vertices corresponding to the views where the $b^{th}$ band exists. Edges are drawn between each node and its direct four neighbors. Isolated nodes are connected to their nearest neighbor. 

The Adjacency is used to compute the inter-view angular unweighted Laplacian as $\mathbf{L}_a = \mathbf{D}_a - \mathbf{A}_a$ with $\mathbf{D}_a$ the degree matrix. $\mathbf{L}_a$ can be diagonalized as:
\begin{equation}
\mathbf{L}_b = \mathbf{U}_a \mathbf{\Gamma} \mathbf{U}_a^\top
\end{equation}
For a specific band number $b$ and super-pixel $k$, the band signal is defined as $b^b_k = \{\hat{f^v_k}(b), \quad  v \subseteq \{1, \cdots, N_v\}\} \in \mathbb{R}^{N_b}$. The angular Graph Transform consists of projecting the signal onto the eigenvectors of $\mathbf{L}_a$ as:
\begin{equation}
c^b_k = \mathbf{U}^\top_ab^b_k
\end{equation}
The inverse angular Graph Transform is then given by
\begin{equation}
b^b_k= \mathbf{U}_a {c}^b_k 
\end{equation}

\subsubsection{Coherence of spatial graph transforms in corresponding super-pixels}
The spatial graphs in the different super-pixels forming one super-ray may not have the same shape. Furthermore, we have observed that for a specific super-ray, when the spatial graph topology in the corresponding super-pixels undergoes a slight change, the basis functions of each spatial graph transform are different and thus incompatible with each others (refer to Figure \ref{fig:not__opt} before optimization), resulting in decreased correlation of the spatial transform coefficients across views. This is shown in the sequel to severely decrease the efficiency of the angular transform. 

Basically, during the diagonalization procedure, the eigenfunctions are only defined up to sign flips for Laplacians having a simple spectrum (if the eigenvalues have a multiplicity of 1, for example connected graphs). Therefore, even having the same shape in two different views, we may end up with two opposite eigen-vectors for a specific eigenvalue during the diagonalization.

Moreover, eigenvectors computed independently on two different shapes (i.e. corresponding to two different Laplacians) can be expected to be reasonably consistent only when the shapes are approximately isometric. Whenever this assumption is violated, it is impossible to expect that the $k^{th}$ eigenvector of a Laplacian $\mathbf{L}_{si}$ in view $i$ will correspond to the $k^{th}$ eigenvector of another Laplacian $\mathbf{L}_{sj}$ in view $j$. If the basis functions do not behave consistently on the corresponding points of the two shapes, the two signals defined on those two Laplacians will be projected onto incompatible basis functions (see Figure \ref{fig:not__opt}), and therefore we cannot guarantee any correlation to be preserved after performing the first spatial graph transform. 

\subsubsection{Coherent spatial graph transform}
In order to overcome those limitations, we consider an approach which aims at finding \textit{coupled} basis functions.

More precisely, suppose that, in a super-ray in a reference view $0$ and a target view $i$, we have two Laplacians $\mathbf{L}_{s_o}$ and $\mathbf{L}_{s_i}$ with size $(n_0 \times n_0)$ and $(n_i \times n_i)$ respectively. They can be diagonalized as: 
\begin{equation}
\begin{split}
&\mathbf{L}_{s_0} = \mathbf{U}_{s_0} \mathbf{\Lambda}_0 \mathbf{U}_{s_0}^\top\\
&\mathbf{L}_{s_i} = \mathbf{U}_{s_i} \mathbf{\Lambda}_i \mathbf{U}_{s_i}^\top
\end{split}
\end{equation}

If the two Laplacians are equal, we make sure that their eigenvectors are compatible with sign flips accordingly. We check the first value of the each eigenvector and flip its sign if the value is negative. \\
In the case where the super-pixel shapes in the sub-aperture images are not isometric, we propose to diagonalize one specific spatial graph Laplacian $\mathbf{L}_{s_0}$ and find $\mathbf{U}_{s_0}$. Then, we search for basis vectors $\hat{\mathbf{U}}_{s_i}$ that approximately diagonalize any other spatial graph Laplacian $\mathbf{L}_{s_i}$ and at the same time preserve correlations after the transform. Inspired by the work of \cite{kovnatsky2013coupled}, we pose the problem as 
\begin{equation}
\begin{split}
 \hat{\mathbf{U}}_{s_i}^* = & \min\limits_{\hat{\mathbf{U}}_{s_i}}\  off(\hat{\mathbf{U}}_{s_i}^\top \mathbf{L}_{s_i} \hat{\mathbf{U}}_{s_i}) + \alpha \left\Vert(\mathbf{F}^\top \mathbf{U}_{s_0} - \mathbf{G}^\top \hat{\mathbf{U}}_{s_i})\right\Vert^2_F ,\\
& \text{s.t. } 
\hat{\mathbf{U}}_{s_i}^\top \hat{\mathbf{U}}_{s_i}  = \mathbf{I}.
\end{split}
\label{eq:iter_opt_1}
\end{equation}
where we seek to minimize the weighted sum of two terms subject to the orthonormality constraint of the computed basis functions $\hat{\mathbf{U}}_{s_i}$. The first term is a diagonalization term that aims at minimizing the energy residing on off-diagonal entries ($off(\mathbf{M}) = \sum_{i\neq j}{m_{ij}})$. The second term aims at enforcing coherence between the two spatial graph transforms and is defined as follows.

\begin{figure}[h]
	\centering
  \includegraphics[width=0.6\textwidth]{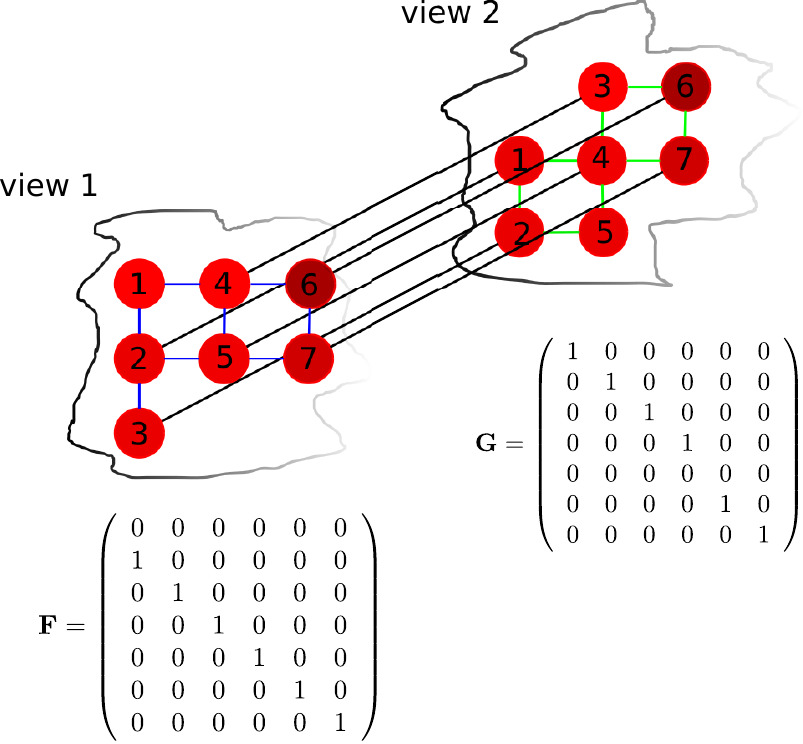}
\caption{Example of correspondence functions $\mathbf{F}$ and $\mathbf{G}$ computed for a small shape-varying super-pixel. The graph nodes are labeled in both graphs following a vertical scan line. In the second view, we have one disappearing node and another appearing one with respect to the first view.}
\label{fig:FGillustration}
\end{figure}

Based on the geometry information we have in hand, we can actually define, \textit{a priori}, a set of correspondences between $\mathbf{L}_{s_0}$ and $\mathbf{L}_{s_i}$. More precisely, we suppose that we have a set of $p$ corresponding functions represented by matrices $\mathbf{F}$ and $\mathbf{G}$ of sizes $(n_0 \times p)$ and $(n_i \times p)$ respectively. An example of $\mathbf{F}$ and $\mathbf{G}$ is shown in figure \ref{fig:FGillustration}.

The basis functions of both Laplacians are supposed to be consistent if the Fourier coefficients of the functions $\mathbf{F}$ and $\mathbf{G}$ on $\mathbf{L}_{s_0}$ and $\mathbf{L}_{s_i}$ are approximately equal i.e. if $\mathbf{F}^\top \mathbf{U}_{s_0} \simeq \mathbf{G}^\top \hat{\mathbf{U}}_{s_i}$. To avoid over-determining the problem, we use the farthest point sampling technique restricting the correspondence points to a maximum of $15$ points.
\begin{figure*}[htp]
  \centering
  \includegraphics[width=0.95\textwidth]{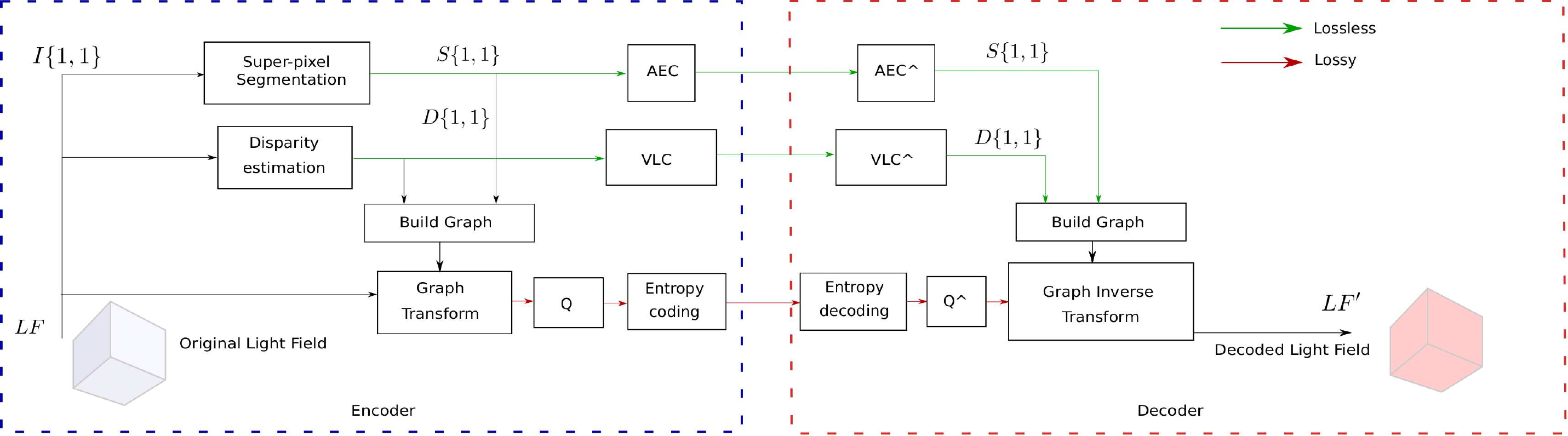}
  \caption{Overview of proposed coding scheme.}
  \label{fig:CodingScheme1}
\end{figure*}

If we parametrize the new basis functions of $\mathbf{L}_{s_i}$ as being a linear combination of the old basis functions, we can write $\hat{\mathbf{U}_{s_i}} = \mathbf{U}_{s_i} \mathbf{B}$ where $\mathbf{B}$ is a matrix of combination coefficients, that plays a role of reflecting and rotating the original basis vectors in $\mathbf{U}_{s_i}$ so that they will align the best way with $\mathbf{U}_{s_0}$ while almost diagonalizing the laplacian $\mathbf{L}_{s_i}$. Using the diagonalizing
property of $\mathbf{U}_{s_i}$, we can re-write Equation (\ref{eq:iter_opt_1}) as 
\begin{equation}
\begin{split}
 \mathbf{B}^* = & \min\limits_{\mathbf{B}}\  off(\mathbf{B}^\top \Lambda_i \mathbf{B}) + \alpha \left\Vert(\mathbf{F}^\top \mathbf{U}_{s_0} - \mathbf{G}^\top \mathbf{U}_{s_i} \mathbf{B})\right\Vert^2_F ,\\
& \text{s.t. } 
\mathbf{B}^\top \mathbf{B} = \mathbf{I},
\end{split}
\label{eq:iter_opt_2}
\end{equation}

It is important to note that the first term of the above problem does not guarantee a preserved increasing order of the eigenfunctions. It is therefore more convenient to use an alternative penalty equal to $\left\Vert \mathbf{B}^\top \mathbf{\Lambda}_i \mathbf{B} - \mathbf{\Lambda}_i \right\Vert^2_F$ that relates not only to the diagonalization property, but also to the distribution of the energies across the basis functions after the optimization.

\begin{equation}
\begin{split}
 \mathbf{B}^* = & \min\limits_{\mathbf{B}}\  \left\Vert \mathbf{B}^\top \mathbf{\Lambda}_i \mathbf{B} - \mathbf{\Lambda}_i \right\Vert^2_F + \alpha \left\Vert(\mathbf{F}^\top \mathbf{U}_{s_0} - \mathbf{G}^\top \mathbf{U}_{s_i} \mathbf{B})\right\Vert^2_F ,\\
& \text{s.t. } 
\mathbf{B}^\top \mathbf{B} = \mathbf{I},
\end{split}
\label{eq:iter_opt_3}
\end{equation}

The problem in Equation (\ref{eq:iter_opt_3}) is a non linear optimization problem with an orthogonality constraint, which can be solved by iterative minimization algorithms. In our case, we used Matlab optimization toolbox (interior point method of the \textit{fmincon} function) to solve it. The gradients of the cost function terms are given in appendix A. 

Since we are dealing with large datasets and a large number of super-rays, it is convenient to use parallel computing to independently compute eigen-basis for the different super-rays. 
Also, in order to reduce the complexity of the problem, we propose to split it into smaller problems that are independent: we pick a small number $k$ of eigenvectors to be optimized at a time. Then, for each disjoint group $l$ of $k$ eigenvectors in $\mathbf{U}_{s_i}$, we formulate a sub-problem by expressing $k$ new eigenvectors as a linear combination of $k$ old eigenvectors. Noticing that $\mathbf{U}_{s_i} = [\widetilde{\mathbf{U}}_{s_{i_1}},\widetilde{\mathbf{U}}_{s_{i_2}} ,... ,\widetilde{\mathbf{U}}_{s_{i_l}}]$ and 
\begin{equation}
\mathbf{\Lambda}_i = \left(\begin{array}{cccc} \widetilde{\mathbf{\Lambda}}_i^1 & 0 & 0 & 0\\
 0 & \widetilde{\mathbf{\Lambda}}_i^2 & 0 & 0\\
 0 & 0 & .. & 0\\
 0 & 0 & 0 & \widetilde{\mathbf{\Lambda}}_i^l \end{array}\right)
\end{equation}
For each group of $k$ eigenvectors, we find $\widetilde{\mathbf{B}_l}$ of size $(k\times k)$ that will minimize the objective function on the subset of eigenvectors.
\begin{equation}
\begin{split}
\widetilde{\mathbf{B}}_l^* = & \min\limits_{\widetilde{\mathbf{B}}_l}\  \left\Vert \widetilde{\mathbf{B}}_l^\top \widetilde{\mathbf{\Lambda}}_y^l \widetilde{\mathbf{B}}_l - \widetilde{\mathbf{\Lambda}}_y^l \right\Vert^2_F+\alpha \left\Vert(\mathbf{F}^\top \widetilde{\mathbf{U}}_{s_{0_l}} - \mathbf{G}^\top \widetilde{\mathbf{U}}_{s_{i_l}} \widetilde{\mathbf{B}}_l)\right\Vert^2_F ,\\
& \text{s.t. } 
\widetilde{\mathbf{B}}_l^\top \widetilde{\mathbf{B}}_l = \mathbf{I},
\end{split}
\label{eq:iter_opt_4}
\end{equation}

At the end of the optimization stage, most of the eigenvectors are thereby compatible across views and the transform will necessarily preserve any correlation already observed between views. An example of the second eigenvector of a super-ray before and after optimization is shown in Figure \ref{fig:not__opt}. While eigenvectors corresponding to higher frequencies are harder to adjust, the low frequency eigenvectors can be easily optimized.  In our application, this is not a big problem since we have a high energy compaction in lower frequency bands, and those are the bands that matter the most for reconstruction. After performing the segmentation and two transforms, most of the energy of the color signal is indeed expected to be concentrated in a very small number of coefficients. In the following section, we aim at exploiting this energy compaction property to efficiently code the redundant information present in the light field using the tools introduced above.

\section{Light Field Coding Scheme}
\label{sec:schemes}
The overall steps of the compression algorithm are shown in Figure \ref{fig:CodingScheme1}. The top left view of the Light Field is separated into uniform regions using the SLIC algorithm to segment the image into super-pixels \cite{achanta2012slic}, and its disparity map is estimated. Using both the segmentation map and the geometry information, we construct consistent super-rays in all views as explained in section \ref{sec:SRconstruction}. 
The non separable and separable transforms described above are then locally applied on each super-ray.
The transformed coefficients are then quantized and encoded to be stored or transmitted. The segmentation map of the reference view and a disparity value per super-ray also need to be transmitted as side information to the decoder.

\begin{figure*}[!htp]
  \centering
  \includegraphics[width=0.95\linewidth]{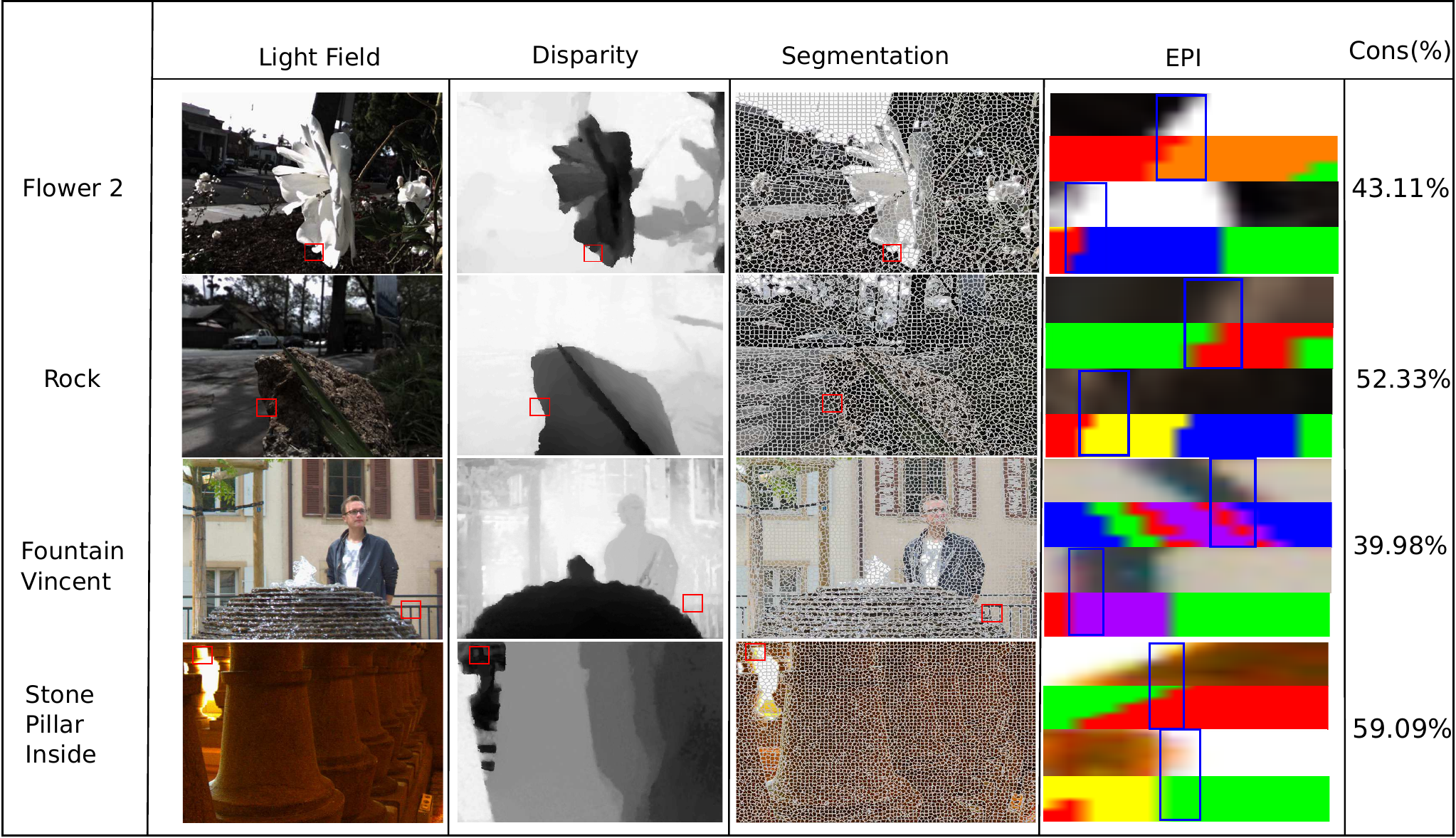}
  \caption{Consistent Super-rays performance:In the first three columns, we have the original top left corner view, its corresponding disparity map and super pixel segmentation using the SLIC algorithm \cite{achanta2012slic} respectively. In the fourth column, we show horizontal and vertical epipolar segments taken both from the 4D light field color and our final labeling in specific regions of the image(the red blocks). We use the prism color map in Matlab for the segmentation, just for illustration purposes.}
  \label{fig:Superrays_perf}
  \vspace{-0.5cm}
\end{figure*}

\subsubsection{Segmentation map and disparity values coding}
The segmentation map of the reference view is encoded using the arithmetic edge coder proposed in \cite{daribo2012arithmetic}. The contours are first represented by differential chaincode \cite{Freeman1961on} and divided into segments. Then, to efficiently encode a sequence of symbols in a segment, \textit{AEC} uses a linear regression model to estimate probabilities, which are subsequently used by the arithmetic coder. Disparity values are encoded using an arithmetic coder. 

\subsubsection{Grouping and transform coefficients coding}
\label{subsubsec:Grouping}
The energy compaction is not the same in all super-rays. This can be explained by the fact, that the segmentation may not well adhere to object boundaries, resulting in high angular frequencies after optimization of the first spatial transform. 

To optimize the coding performance, we divide the set of super-rays into four classes, where each class is defined according to an energy compaction criterion. 

First, we learn a scanning order. More precisely, at the end of the two graph transform stages, coefficients are grouped into a three-dimensional array $\mathbf{R}$ where $\mathbf{R}(i_{SR},i_{bd},v)$ is the $v^{th}$ transformed coefficient of the band $i_{bd}$ for the super-ray $i_{SR}$. Using the observations on all the super-rays in some training datasets (\textit{Flower1},\textit{Friends}), we can find the best ordering for scanning and quantization. We sort the variances of coefficients with enough observations in decreasing order and we follow this decreasing order during the scanning process.

Then, each super-ray with $N$ coefficients belongs to class $i$ if the mean energy per high frequency coefficient is less than $1$, where the high frequency coefficients are the last $round(N \times i/4)$ coefficients following the scanning order of the super-rays coefficients defined previously. 
We start by finding the super-rays in the first class than remove them from the search space before finding the other classes, and idem for the following steps. 
We code a flag with an arithmetic coder to gives the information of the class of super-rays to the decoder side. In class $i$, the last $round(N \times i/4)$ coefficients of each super-ray are discarded. The rest of the coefficients are grouped into $32$ uniform groups. The quantization step sizes in groups are defined with a rate-distortion optimization taking into account a big number of observed coefficients. At the end of this stage, for each class, each group is coded using the Context Adaptive Binary Arithmetic Coder (\textit{CABAC}) from the HEVC H.265 reference coder.

\section{Experimental analysis}
For performance evaluation, we consider real light fields captured by plenoptic cameras from the datasets used in \cite{kalantari2016learning} and \cite{viola2018graph}.
We consider the $8 \times 8$ central sub-aperture images cropped to  $364 \times 524$ in \cite{kalantari2016learning}, and $9 \times 9$ cropped to $432 \times 624$ from \cite{viola2018graph} in order to avoid the strong vignetting and distortion problems on the views at the periphery of the light field. The disparity map of the top left view of each light field has been estimated using the method in \cite{jiang2018depth}. The estimated disparity map is used to construct super-rays as described in Section \ref{sec:SRconstruction}. 

\subsection{Assessment of the proposed super-ray construction method}
In this section, we assess how the proposed super-ray construction method deals with occluded and dis-occluded parts, and to which extent the super-rays are consistent despite uncertainty on the disparity information. Figure \ref{fig:Superrays_perf} shows examples of super-rays obtained with different real light fields captured by a Lytro Ilum camera (\textit{Flower 2}, \textit{Rock} used in \cite{kalantari2016learning}, and \textit{FountainVincent}, \textit{StonePillarInside} used in \cite{viola2018graph}).
In the first three columns, we have the original top left corner view, its corresponding disparity map and super pixel segmentation using the SLIC algorithm \cite{achanta2012slic} respectively. In the fourth column, we show horizontal and vertical epipolar segments taken both from the 4D light field color information and our final segmentation in specific regions of the image (the red blocks). 
We can see that we are following well the object borders, especially when the disparity map is reliable. Also, we have always attained a high percentage of coherent super-rays across views (higher than $40\%$ as measured with Cons($\%$) in the fifth column). More precisely, Cons($\%$) gives the percentage of coherent super-rays: A super-ray is coherent when it is made of super pixels having the same shape in all the views, with or without a displacement.

At the end of this segmentation stage, we end up with a segmentation map with consistent super rays in flat objects and shape-varying super-rays mainly on the borders. 

\subsection{Analysis of proposed graph based optimized transforms}
\begin{figure}[h]
  \centering
  \includegraphics[width=0.45\textwidth]{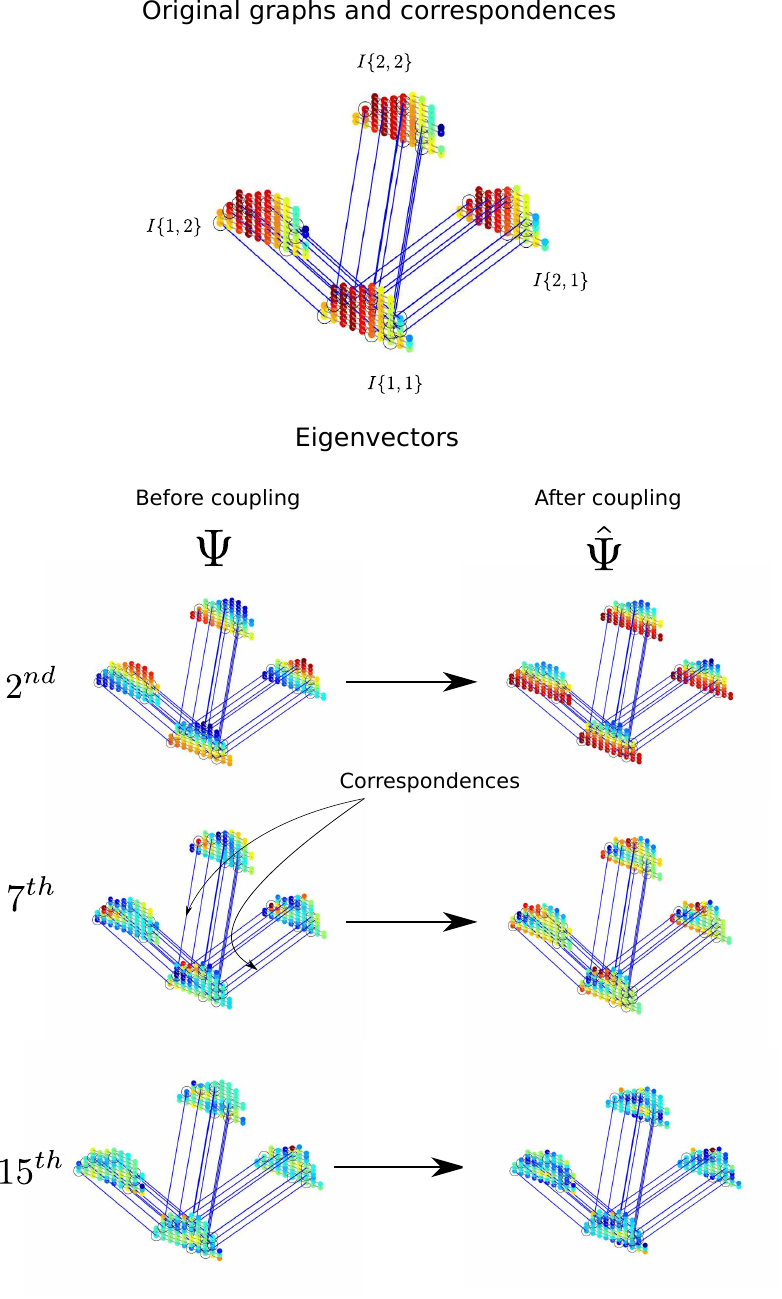}
  \caption{Illustration of the output of the optimization process for a super-ray in 4 views. The first row corresponds to a super-ray accross four views of the light field. The signal on the vertices correspond to the color values lying on super-pixels corresponding to the same super-ray and the blue lines denote the correspondences. The second to fourth rows are illustrations of basis functions before and after optimization. The signals on the vertices are the eigenvectors values.}
  \label{fig:AfterOptimization_Experimental}
\end{figure}

In this section, we analyze the performance of our optimization process described in section \ref{sec:CSGT} and its effect on the transform coding efficiency. In all the experiments, for each super-ray we find the super-pixel $\mathbf{L}_{s_o}$ that is on the top-left most of the light field, and fix it as reference for the coupling process. We therefore optimize the maximum number of eigenvectors defined as $floor(\frac{n_{0}}{10})\times10$ with $n_0$ being the number of pixels in the reference super-pixel. An example of input and output of the coupling process for a shape-varying super-ray is illustrated in Figure \ref{fig:AfterOptimization_Experimental}.
We see that the consistency of eigenvectors in the different graphs is much better after our optimization. If we project the light field signal residing in the super-ray on the optimized coupled eigenvectors, the inter-view correlation is better preserved compared to the non optimized eigenvectors.

\begin{figure*}[!htp]
  \centering
  \includegraphics[width=0.32\linewidth]{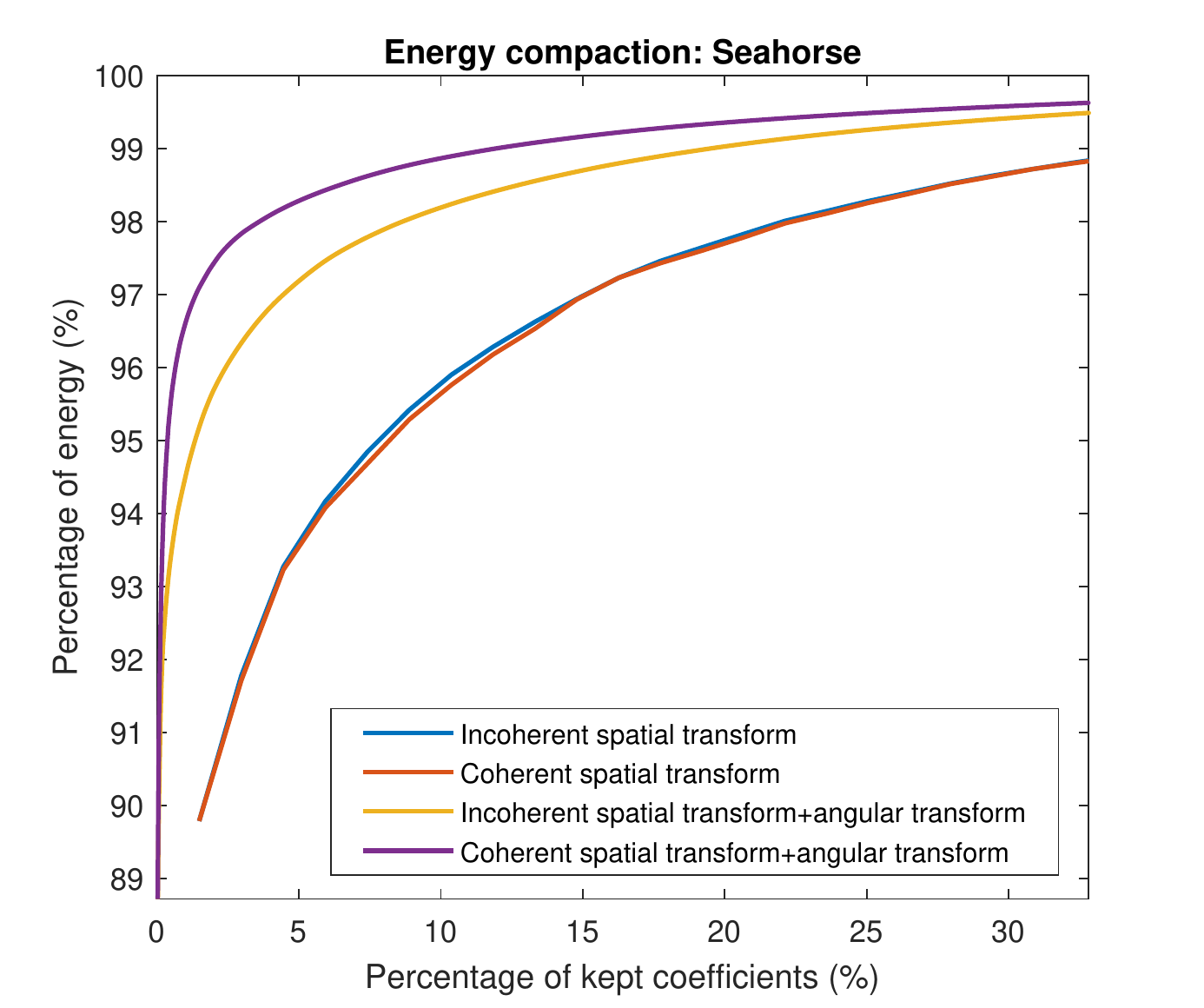}
  \includegraphics[width=0.32\linewidth]{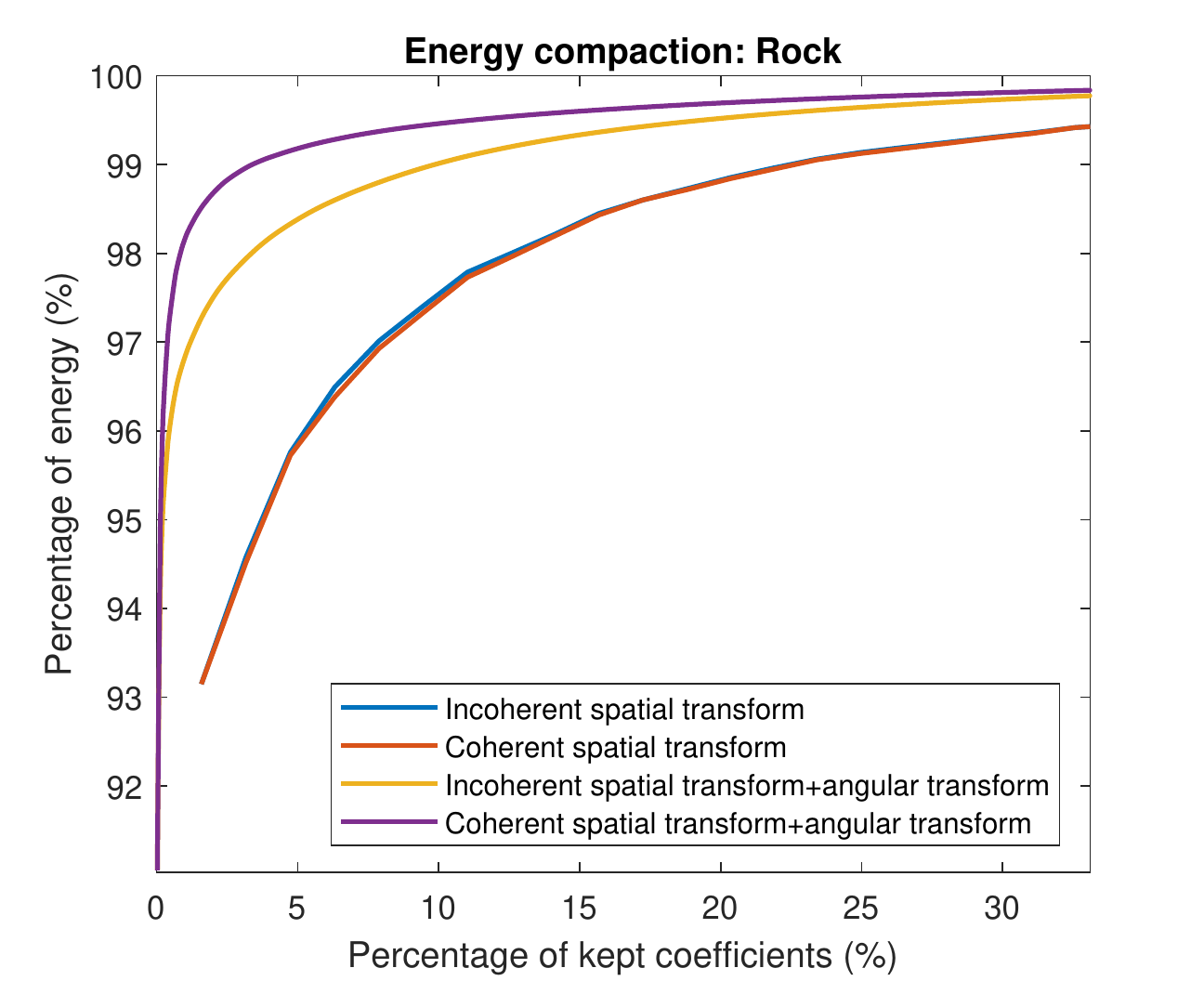}
   \includegraphics[width=0.32\linewidth]{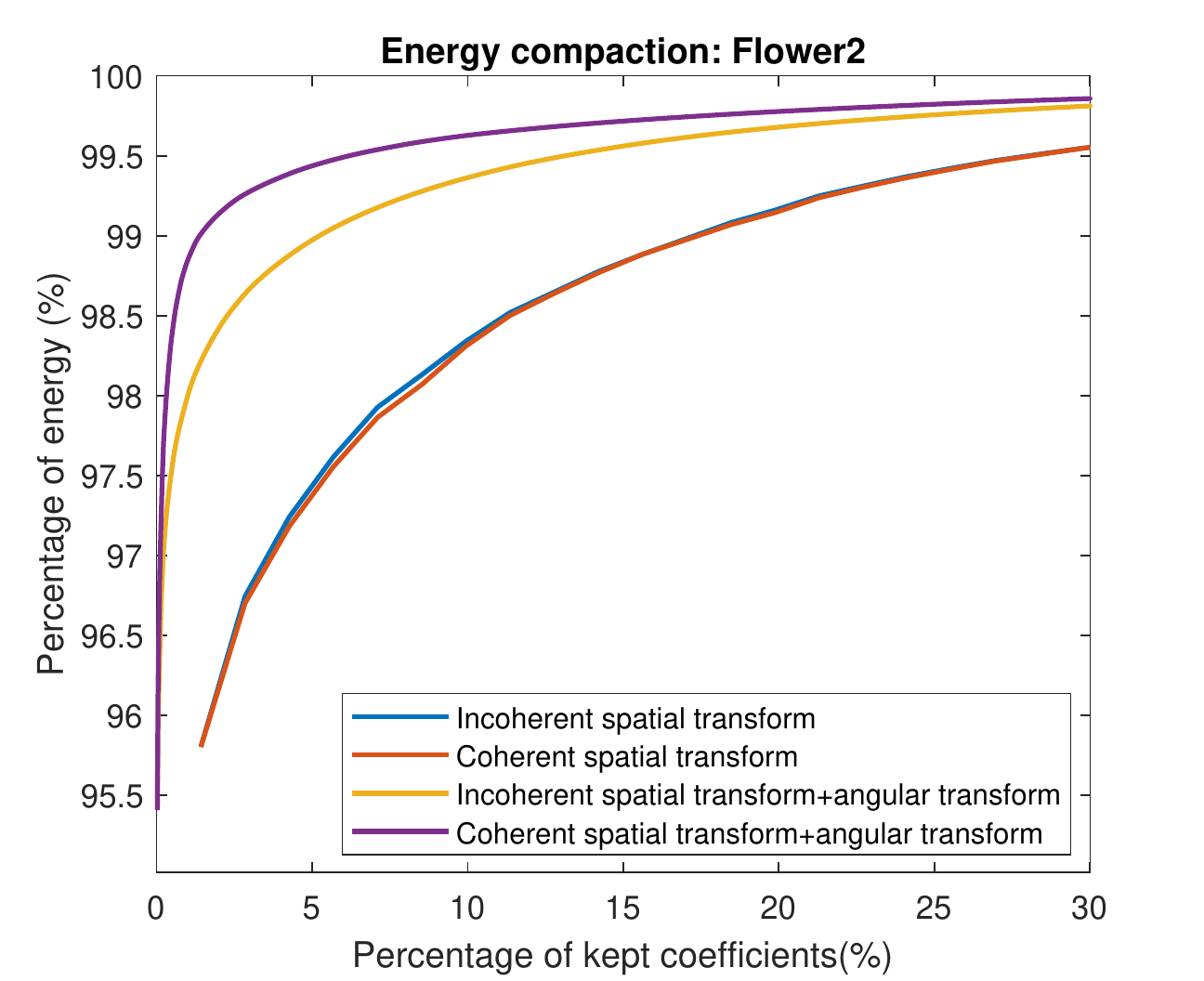}
  \\
  \includegraphics[width=0.32\linewidth]{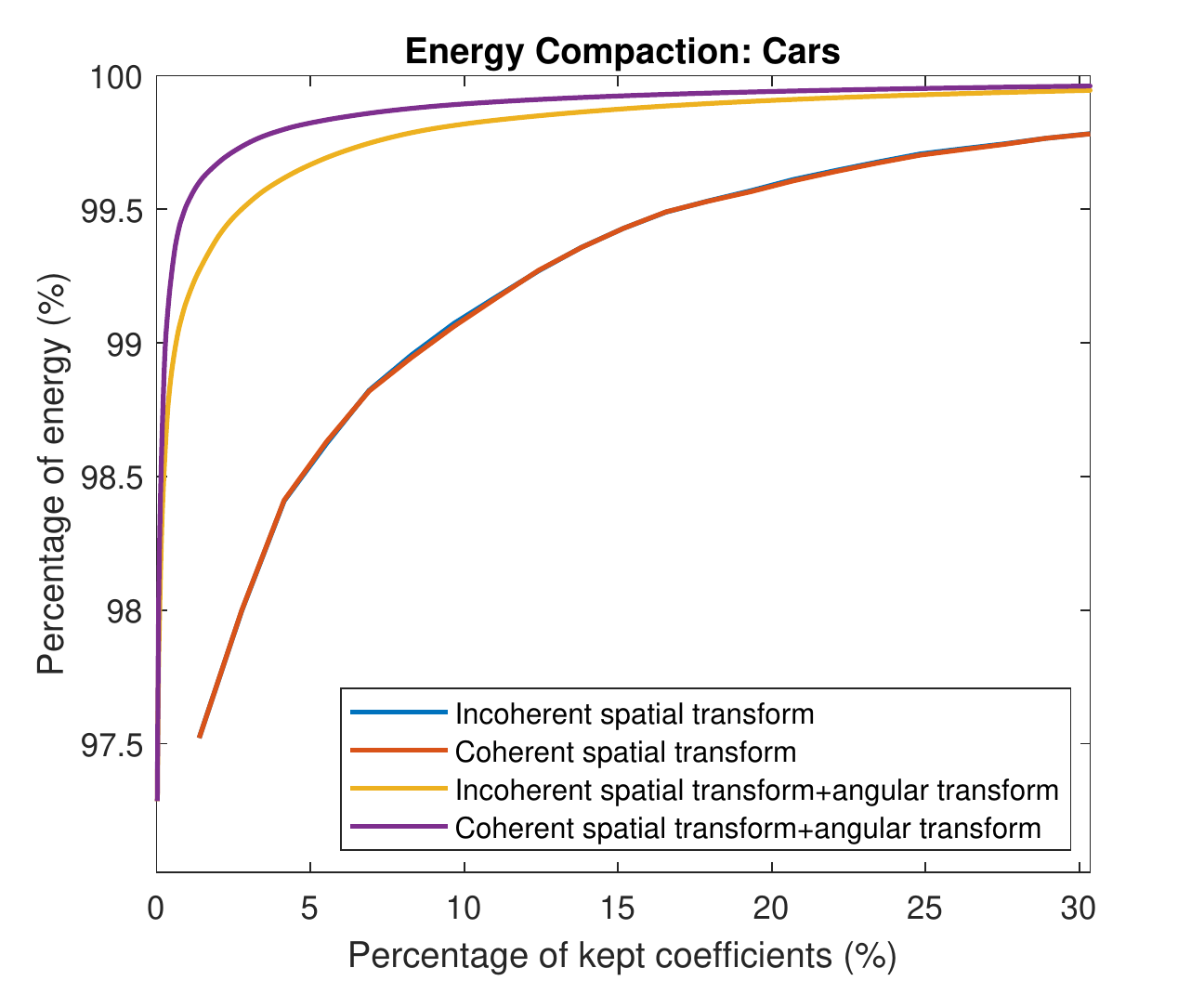}
  \includegraphics[width=0.32\linewidth]{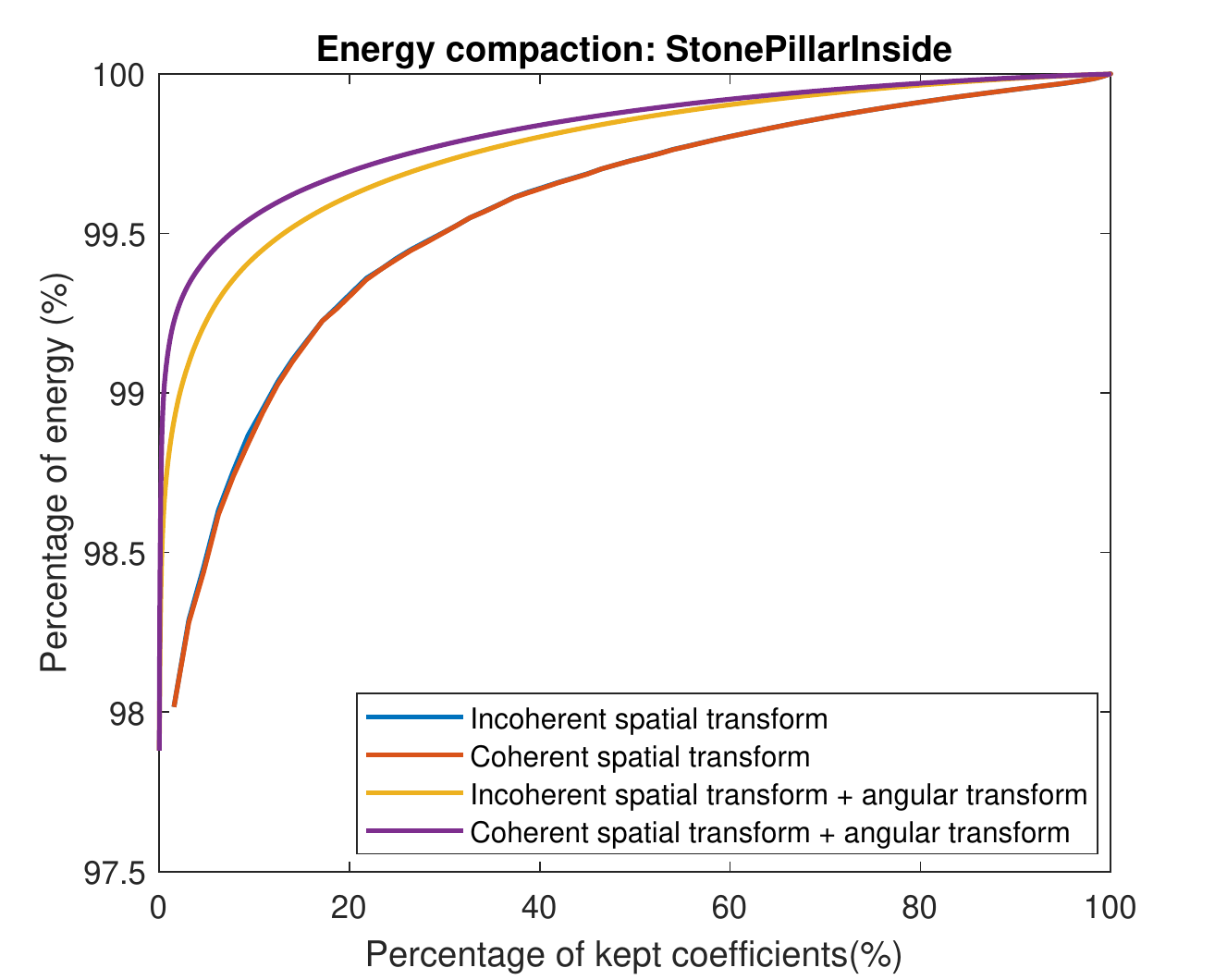}
  \includegraphics[width=0.32\linewidth]{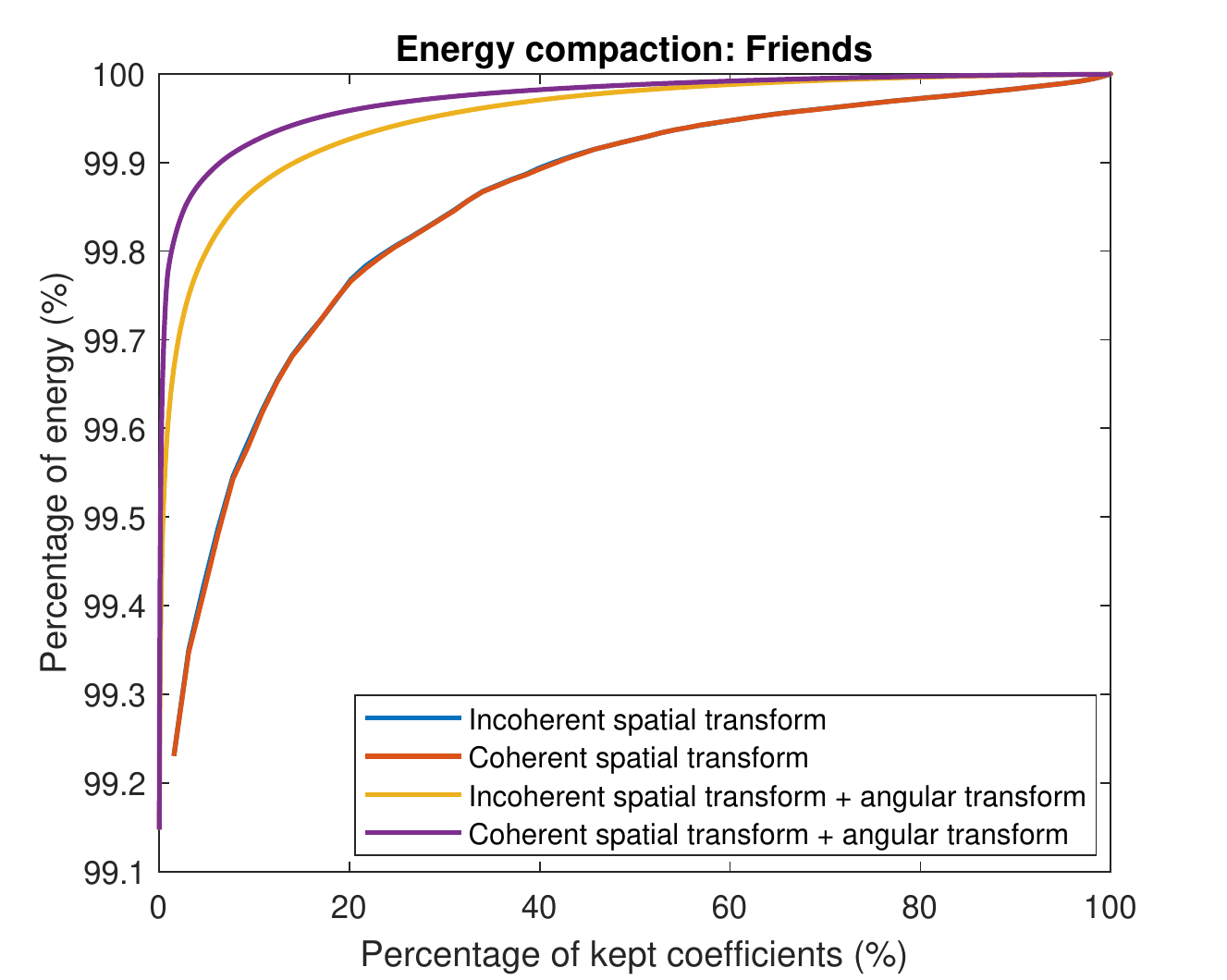}
  \caption{Energy compaction with or without optimization of the first spatial transform for four datasets (Seahorse, Rock,Flower2 and Cars) from the dataset used in \cite{kalantari2016learning} and two others (Friends and StonePillarsInside) taken from the datasets in \cite{viola2018graph}.}
  \label{fig:EnergyCompactionExample}
  \vspace{-0.5cm}
\end{figure*}

\subsubsection{Energy Compaction of the spatial transform}
Figure \ref{fig:EnergyCompactionExample} shows the energy compaction observed in the spatial transform domain, then in the spatio-angular transform domain, \textit{i.e.} after performing the first spatial transform and after performing both spatial and angular transforms on the color signal of the light fields. The energy compaction is computed for both optimized and non optimized cases. It denotes the percentage of energy if we keep some of the coefficients and discard others. For the spatial transform, we gather the transform coefficients of all super-pixels, and then we scan them following the intuitive order increasing order of the Laplacian eingevalues to compute the compaction. For the spatio-angular compaction, we follow the learned sub-optimal scanning order using different observations from the different datasets as explained in section \ref{subsubsec:Grouping}. 

\begin{figure}[h]
  \centering
  \includegraphics[width=0.7\textwidth]{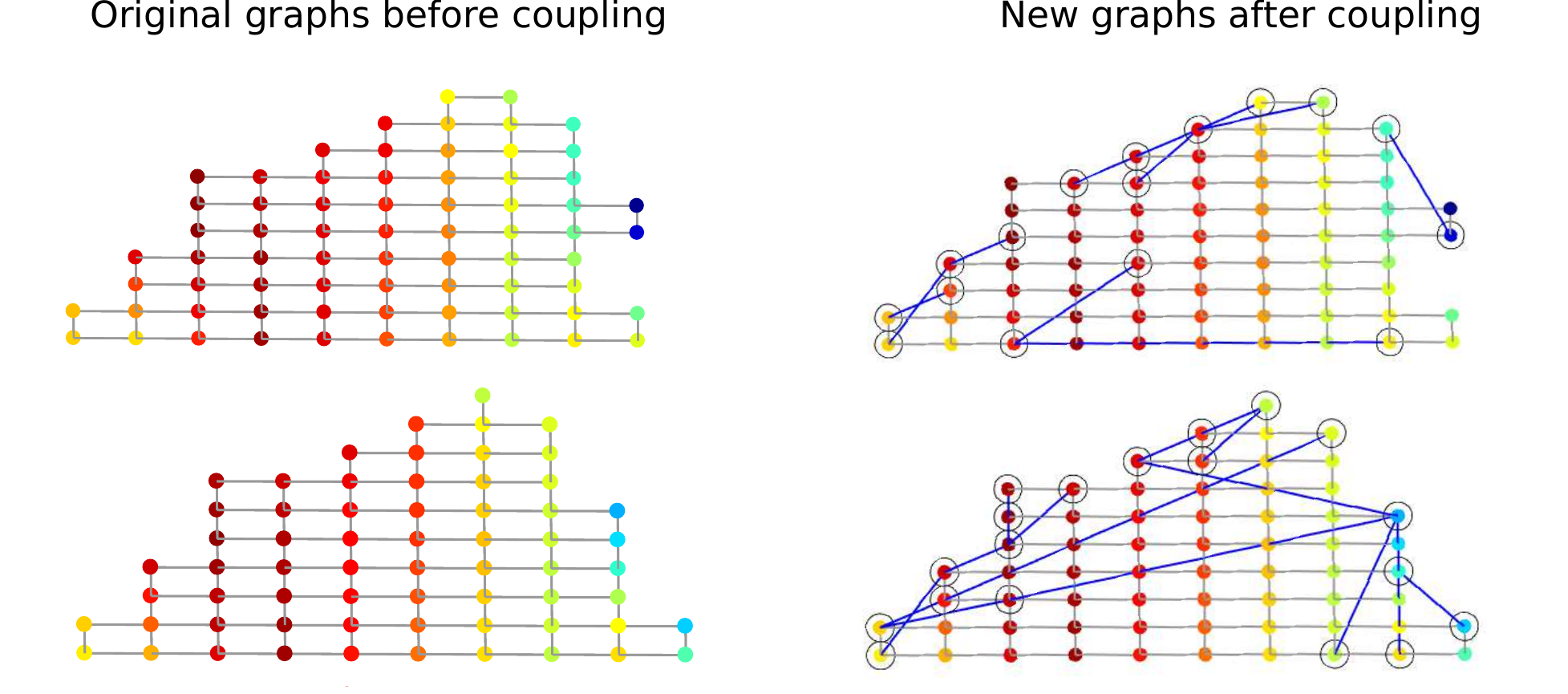}
  \caption{Image showing the old graphs before coupling and the new graphs after optimization. New edges with absolute weight values larger than 0.04 are shown as blue lines connecting highlighted nodes.}
  \label{fig:NewGraphs}
\end{figure}

If we compare the energy compaction of the spatial transforms only (red and blue curves) for different datasets, we observe that we may loose in terms of energy compaction for some datasets after optimization. 
In order to explain such loss, we analyze how the graphs are varying under the new basis functions after optimization. An example is shown in Figure \ref{fig:NewGraphs} where edges between highlighted nodes are added implicitly in the graph after coupling. The new underlying Laplacian is computed as $\hat{\mathbf{L}}_{s_i} = \hat{\mathbf{U}}_{s_i} \mathbf{\Lambda}_i \hat{\mathbf{U}}_{s_i}^T$. 

The underlying assumption behind the optimization procedure is that the signal can be modeled by a modified Gaussian distribution (Gaussian Markov Random Field) with a modified precision matrix which is equivalent to the new Laplacian matrix with some added small weights. Since this procedure is modifying the original graph structure, it may, in some cases, bring some high frequencies.

\subsubsection{Correlation and Energy Compaction after angular transform}
The gain in compaction after the spatio-angular transform is clear in Figure \ref{fig:EnergyCompactionExample} when we perform the optimization. This is due to the fact that we are able to preserve angular correlations after the spatial transform, which will be subsequently exploited by the angular transform. 

In order to assess the performance of our coupling process in preserving the correlation, we draw in Figure \ref{fig:Compact_perf}, the correlation matrices and the covariance matrices for some bands after the first transform 
with shape-varying super-rays. If we restrict our attention to the first column, We see that after the first transform that is not optimized, we have uncorrelated transform coefficients due to the perturbation of eigenvectors computed on super-pixels having slightly different shapes. This problem is almost resolved with our coupling procedure in the second column, where we can observe more correlation between the coefficients of the same band in neighboring views. 
{Furthermore, the logarithm of the variances (values lying on the diagonal in the covariance matrices) being higher in the first low frequency bands and decreasing when moving further from the DC, shows the energy compaction of the first transform. As for the values of the off-diagonal elements of the covariance matrices, they show how correlated are the transformed coefficients after the first transform inside the views. If we observe the off-diagonal values and compare them with or without optimization, we find out that the optimization performs better for low frequencies than for high frequencies and is therefore more able to retrieve coherent basis functions. 

After the second angular transform per band, for both cases with or without optimization, we compute the logarithm of coefficients' variances after the second transform and illustrate it in the third row where the x-axis and y-axis correspond to the band number and the view number respectively. A compaction of the energy in fewer coefficients is observed in the optimized case compared to the non-optimized case, especially when we focus on the top-left region. Some inter-view high frequencies are sometimes still there and might be due to the presence of some super-rays are made of super-pixels that adhere well to borders in some views while not adhering in some others due to disparity rounding effects.

\begin{figure*}[!htp]
  \centering
  \includegraphics[width=0.95\linewidth]{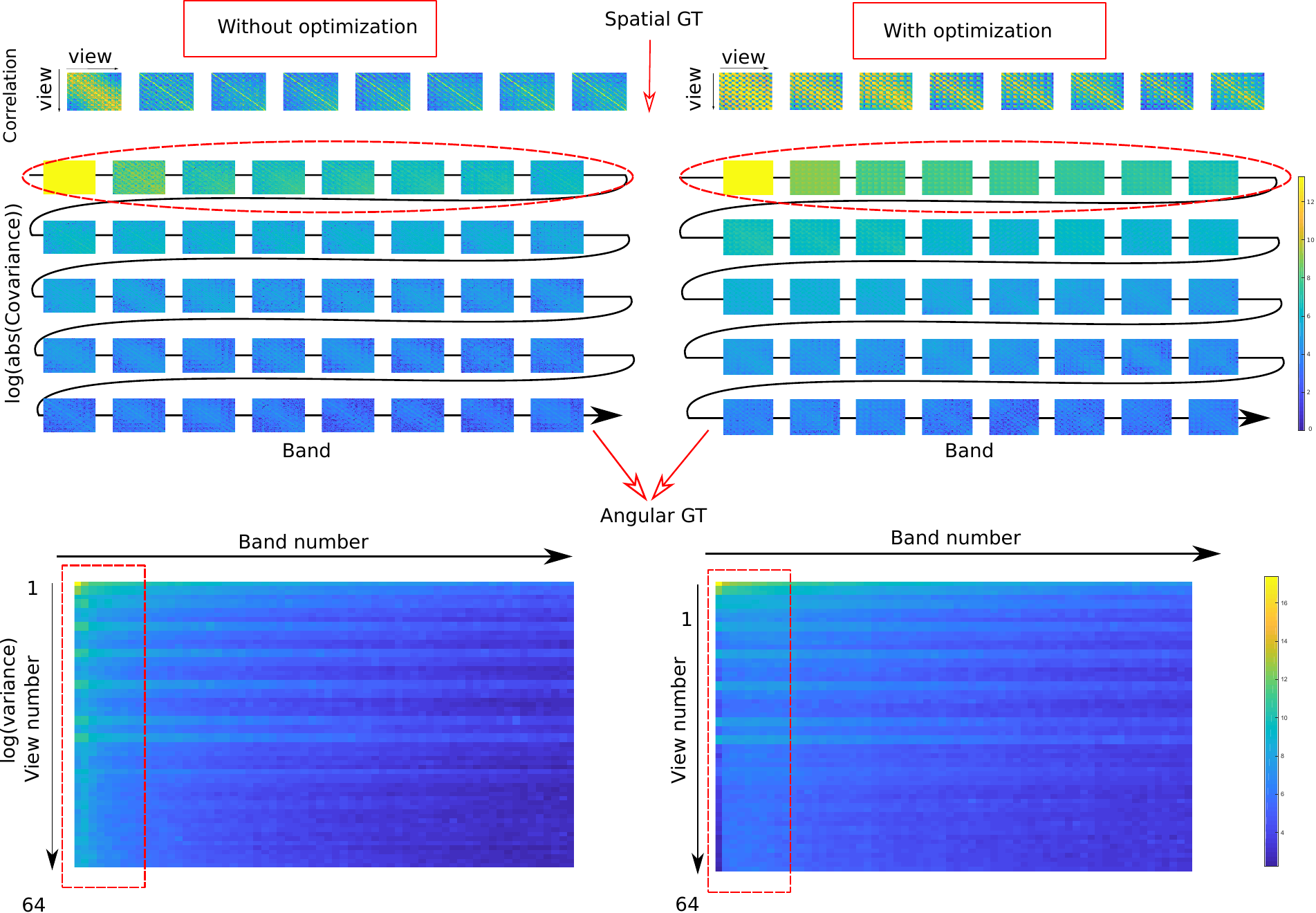}
  \caption{Advantage of our optimization in terms of energy compaction. The three rows correspond to (1) correlation matrices of the spatial transformed coefficients of the first ten bands, (2) the log of the absolute value of the covariance matrices of the 64 first bands of the spatial transformed coefficients, and (3) the logarithm of the variance of the coefficients after the angular transform, respectively. The two columns show the two cases: without or with our optimization.}
  \label{fig:Compact_perf}
\end{figure*}

\subsubsection{Impact of disparity errors}
When the disparity information is not reliable, dis-occluded pixels may be clustered with a wrong super-ray, resulting in high frequencies, hence poor energy compaction, after the spatial transforms in those specific regions.  As explained before, we overcome this problem by dividing the super-rays into classes. 

\subsubsection{Impact of super-rays size}
The size of super-rays may have an impact on the rate distortion performance especially when the disparity information is reliable and there is a lot of homogeneous objects. If we have large objects, we might want to merge some small super-rays which makes a non separable graph transform practically unfeasible. Here comes the advantage of an optimized separable graph transform where one can define the number of eigenvectors to be optimized depending on the homogeneity of the shape-varying super-rays inside the views. In this case, the segmentation and disparity costs will more likely drop also since we also have less contours and values to code. 

In our experiments, however, we use a uniform segmentation into super-pixels. We fix the number of super-rays to $2800$ for the light fields in \cite{kalantari2016learning}, and $4000$ for the light fields in \cite{viola2018graph}.
We have observed that when we have a small number of super-rays, the disparity errors may have an impact on the compensation and therefore result in a decreased PSNR-Rate performance. On the other hand, having a very large number of super-rays increases the rate needed for segmentation and limits the dimension of each super-ray, resulting in a smaller benefit in terms of de-correlation of the proposed spatio-angular transform. 

\subsection{Comparative assessment}
We assess the compression performance obtained with our graph based transform coding schemes against two schemes: direct encoding of the views as a video sequence following a lozenge order (HEVC lozenge) \cite{rizkallah2016impact}, and using the JPEG Pleno VM 1.1 software used as anchor in \cite{viola2018graph}.

In the simulations, the basic configuration files of JPEG Pleno VM have been used with small changes in order to be applied on $9\times 9$ views. For HEVC-lozenge, the base QPs are set to $20$, $26$, $32$, $38$ and a GOP of $4$ is used. The HEVC version used in the tests is HM-16.10.

\begin{figure*}[!htp]
  \centering
  \includegraphics[width=0.32\linewidth]{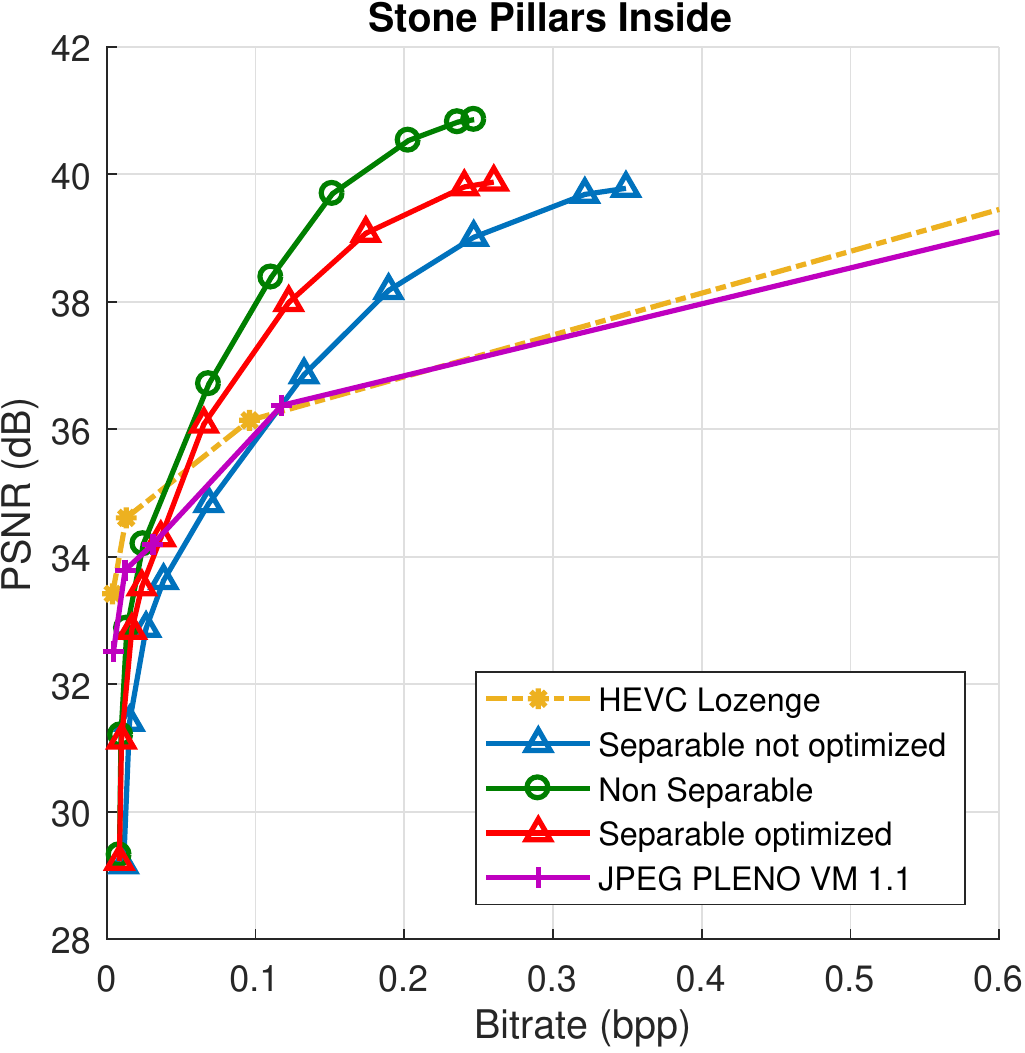}
  \includegraphics[width=0.32\linewidth]{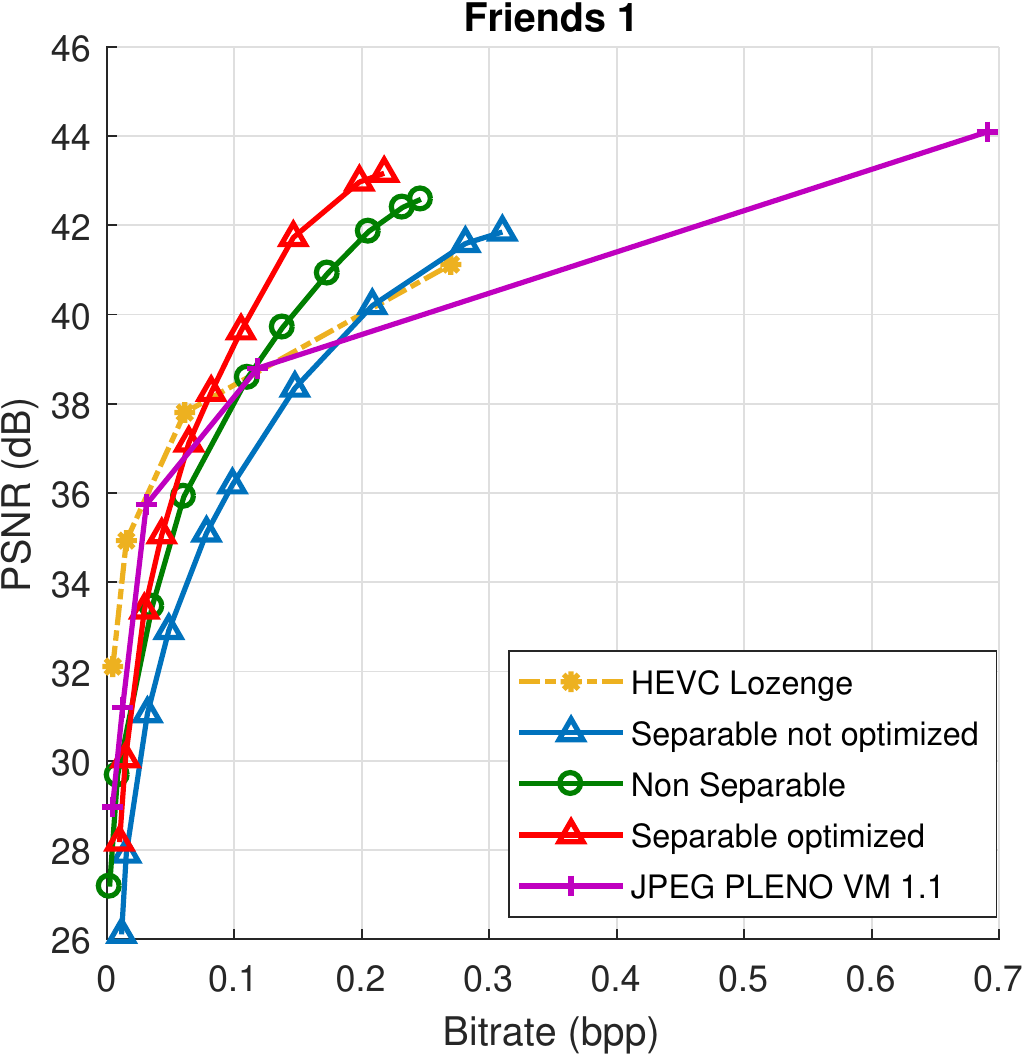}
   \includegraphics[width=0.32\linewidth]{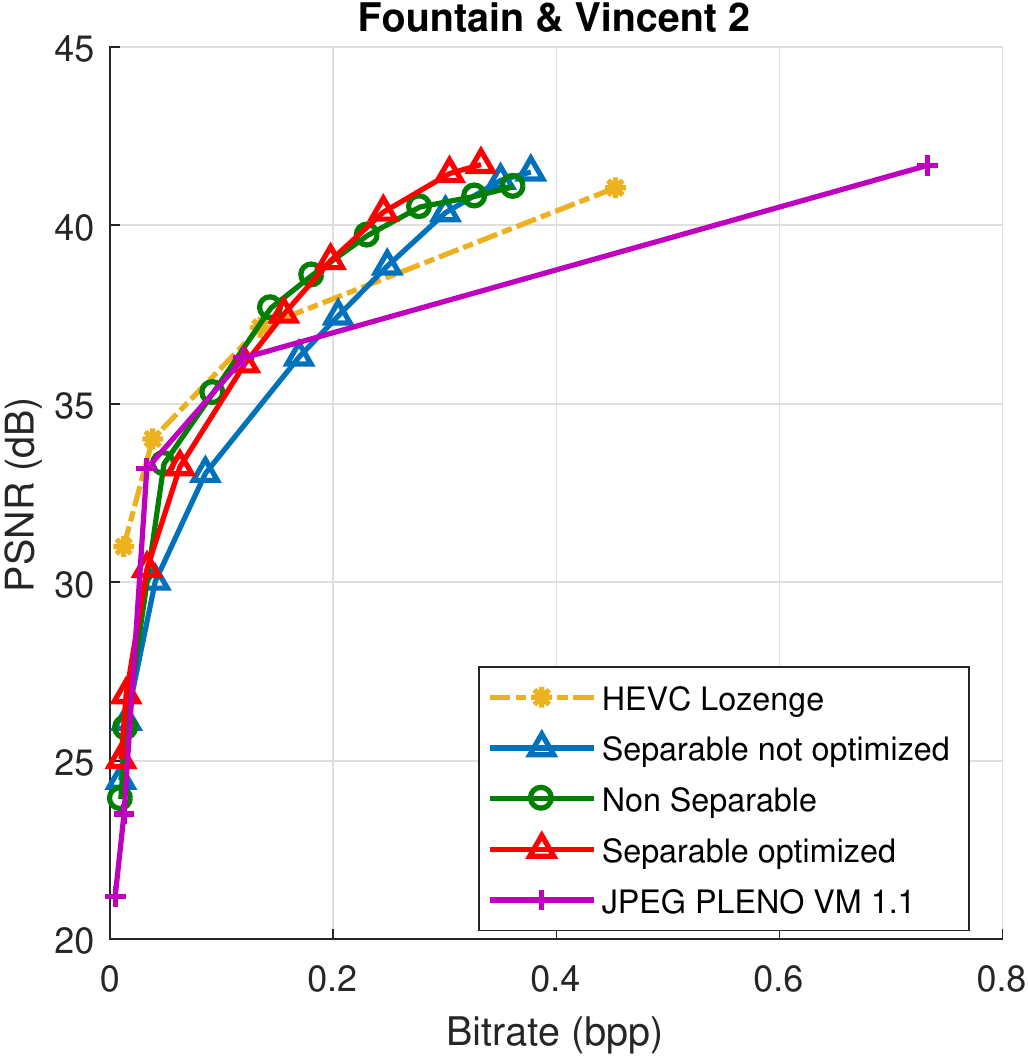}
  \caption{Rate distortion performance of our graph based coding schemes (Non separable, not optimized and optimized separable graph transforms) compared to HEVC lozenge and JPEG Pleno VM 1.1 for the $9\times9$ light fields used in the ICIP 2017 Grand Challenge  \cite{viola2018graph} following the common test conditions.}
  \label{fig:RDcurves_icip}
\end{figure*}

In Figure \ref{fig:RDcurves_icip}, our coding scheme based on both non separable and separable graph transforms is investigated against HEVC-lozenge and JPEG pleno  1.1 for three light fields with $9\times 9$ views, from the ICIP 2017 Grand Challenge \cite{viola2018graph}. Further experiments are also depicted in Figure \ref{fig:RDcurves_kalantari} for $8\times 8$ light fields. For the separable case, we compare the optimized and the non optimized graph transform. In Table \ref{table:rateallocation}, we restrict our attention to the optimized separable graph based transform case that we denote by opt-separable GBT scheme that can be applied no matter how big the super-rays are. It shows the rate allocation of our method, at low and high bitrates, for the different light fields.

We can observe that, for most of the light fields used in our tests, the non separable graph transform yields a better rate-distortion performance compared to the separable case for a fixed number of super-rays. While the non optimized graph transform fails to compact the energy of the light field, the optimized graph transform is performing better and sometimes almost catches the non separable case. One major advantage of the separable optimized case is that it can be applied on super-rays of large dimensions without facing the basis functions computational complexity issue of the non separable case. Furthermore, the number of eigenvectors to be optimized can be defined by the encoder and does not have to be necessarily large.

\begin{figure*}[!htp]
  \centering
  \includegraphics[width=0.24\linewidth]{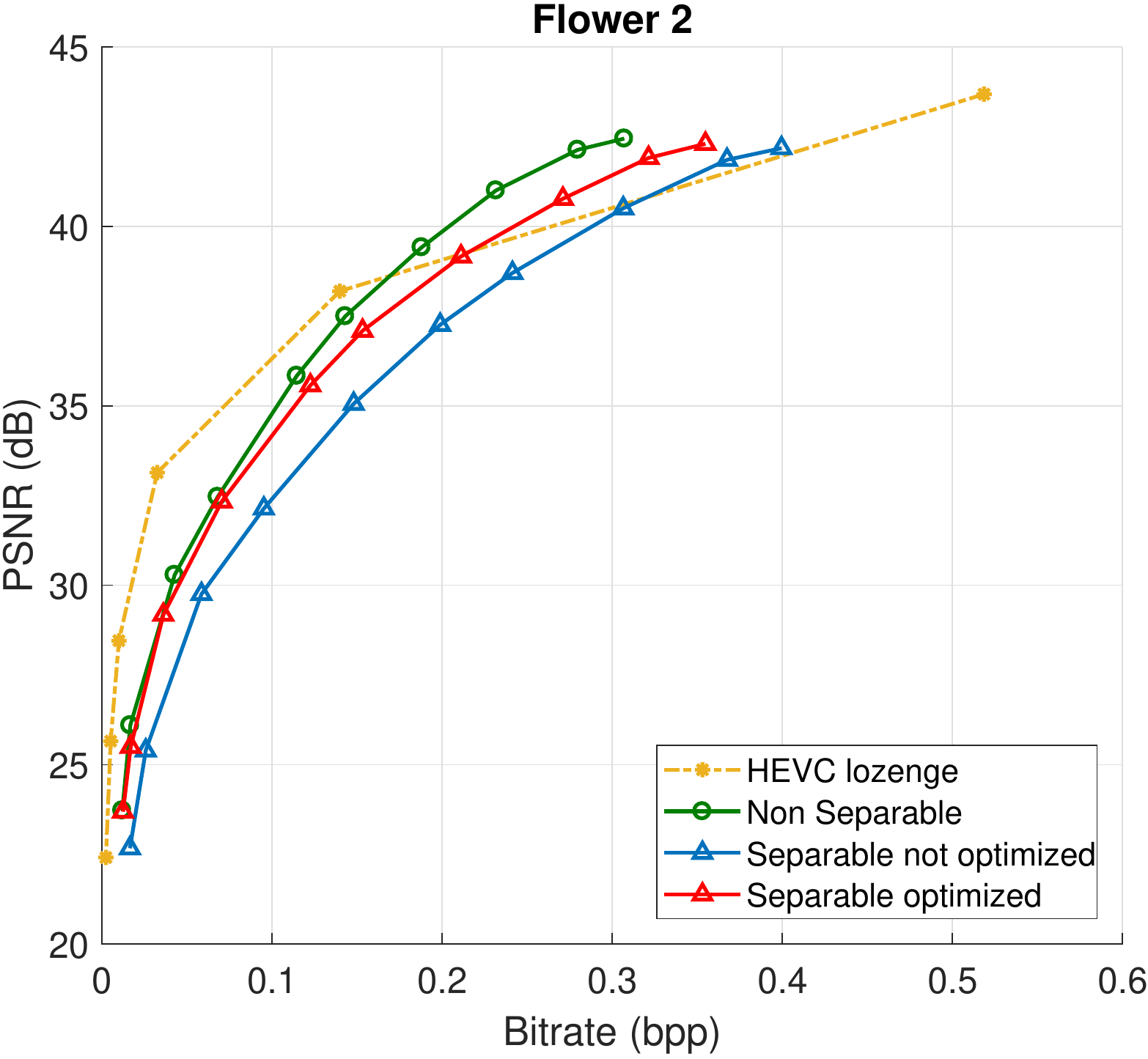}
  \includegraphics[width=0.24\linewidth]{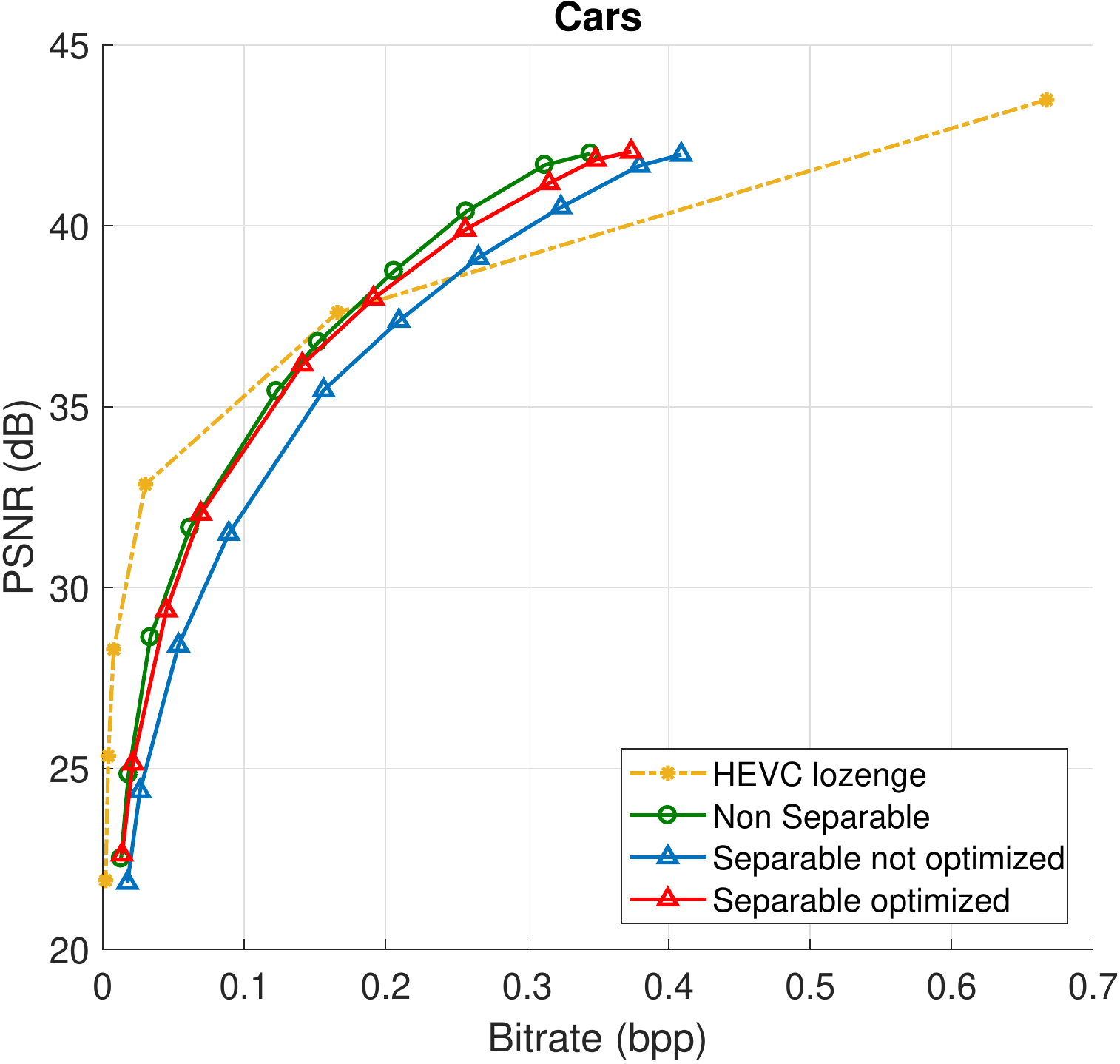}
  \includegraphics[width=0.24\linewidth]{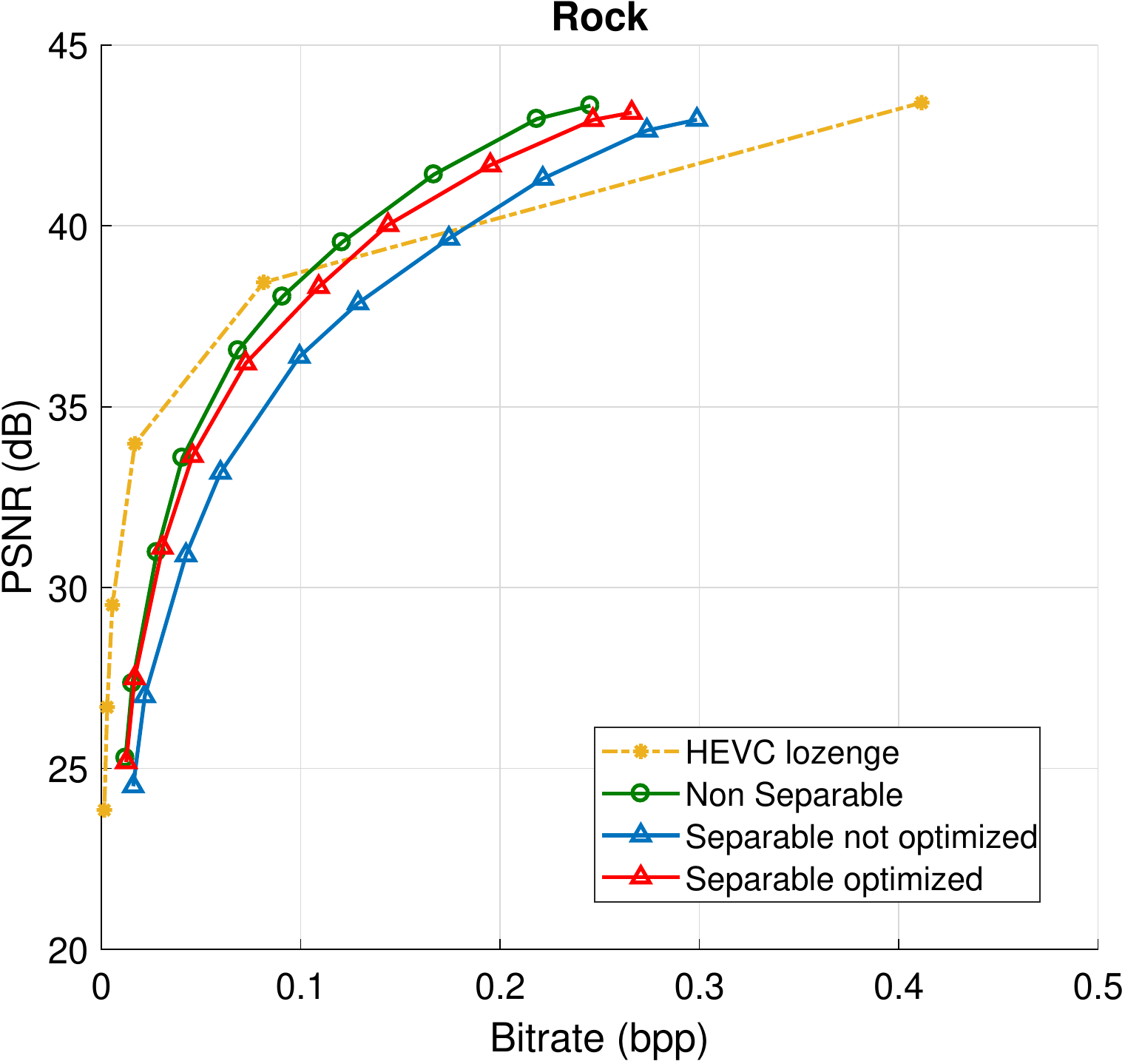} \includegraphics[width=0.24\linewidth]{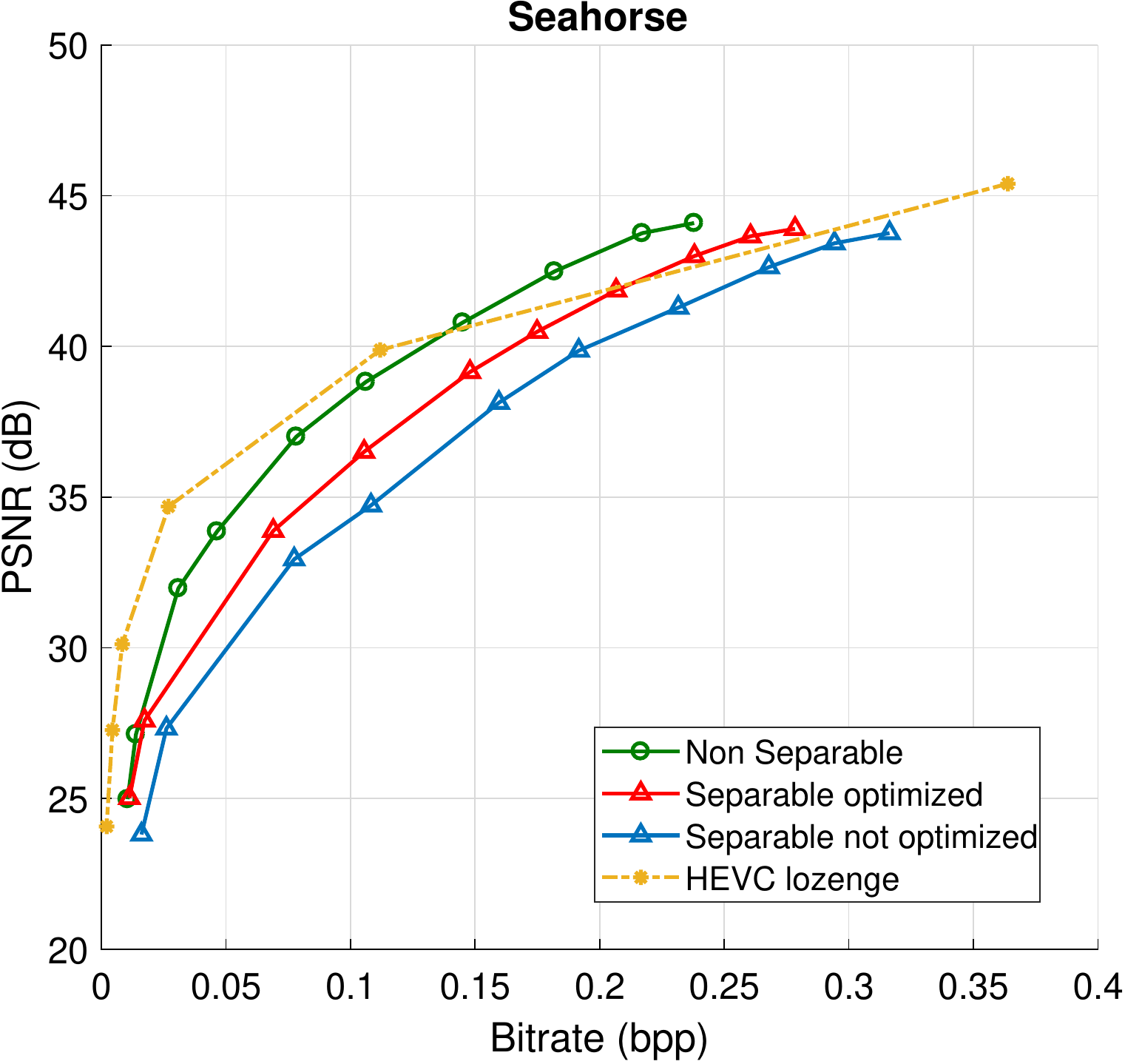}
  \caption{Rate distortion performance of our graph based coding schemes (Non separable, not optimized and optimized separable graph transforms) compared to HEVC lozenge for the $8\times8$ light fields of \cite{kalantari2016learning}. }
  \label{fig:RDcurves_kalantari}
\end{figure*}

\begin{table*}[ht!]
	\centering
	\begin{tabular}{l||cccc}
		\hline
		\multirow{2}{*}{Light Field} & \multicolumn{4}{c}{Rate allocation(in $\%$) for the opt-separable GBT scheme}\\ \cline{2-5} & \textit{Overall bitrate}  & \textit{Segmentation} & \textit{Disparity} & \textit{Coefficients}\\
		\hline
		\multirow{2}{*}{Cars $\left( 364 \times 524 \right)$} & $0.2563$ bpp (PSNR = $42.24$dB) & $2.69\%$ & $0.55\%$ & $96.76\%$ \\ &
		$0.0212$ bpp (PSNR = $25.23$dB) & $32.55\%$ & $6.60\%$ & $60.85\%$\\ \cline{2-5}\\
		\multirow{2}{*}{Flower2 $\left( 364 \times 524 \right)$} & $0.2710$ bpp (PSNR = $40.77$dB) & $2.69\%$ & $0.55\%$ & $96.76\%$ \\ &
		$0.0362$ bpp (PSNR = $29.18$dB) & $20.17\%$ & $4.14\%$ & $75.69\%$\\ \cline{2-5}\\
		\multirow{2}{*}{Rock $\left( 364 \times 524 \right)$} & $0.1951$ bpp (PSNR = $41.68$dB) & $4.00\%$ & $0.82\%$ & $95.18\%$ \\ &
		$0.0306$ bpp (PSNR = $31.10$dB) & $25.49\%$ & $5.23\%$ & $69.28\%$\\ \cline{2-5}\\
		\multirow{2}{*}{Seahorse $\left( 364 \times 524 \right)$} & $0.2302$ bpp (PSNR = $42.99$dB)& $2.65\%$ & $0.74\%$ & $96.61\%$ \\ &
		$0.0612$ bpp (PSNR = $33.88$dB) & $9.97\%$ & $2.78\%$ & $87.25\%$\\
		\hline
		\hline
		\multirow{2}{*}{Friends $\left( 432 \times 624 \right)$} & $0.1464$ bpp (PSNR = $41.73$dB)& $3.89\%$ & $0.10\%$ & $96.01\%$ \\ &
		$0.0294$ bpp (PSNR = $33.38$dB) & $19.39\%$ & $5.10\%$ & $75.51\%$\\ \cline{2-5}\\
		\multirow{2}{*}{StonePillarInside $\left( 432 \times 624 \right)$} & $0.2204$ bpp (PSNR = $39.07$dB) & $2.59\%$ & $0.54\%$ & $96.87\%$ \\ &
		$0.0212$ bpp (PSNR = $32.85$dB) & $26.89\%$ & $5.66\%$ & $67.45\%$\\ \cline{2-5}\\
		\multirow{2}{*}{FountainVincent $\left( 432 \times 624 \right)$} & $0.2448$ bpp (PSNR = $40.37$dB) & $2.12\%$ & $0.57\%$ & $97.31\%$ \\ &
		$0.0330$ bpp (PSNR = $30.38$dB) & $15.76\%$ & $4.24\%$ & $80.00\%$
	\end{tabular}
	\caption{Rate allocation performed by the proposed coding scheme with the optimized separable graph transform. The rate is divided into three parts used for coding the segmentation, disparity and transform coefficients.}
	\label{table:rateallocation}
\end{table*}
Moreover, we can observe a better performance of our method at high bitrate compared to JPEG Pleno VM 1.1 and HEVC lozenge. At low bitrate, the prediction in the HEVC and JPEG Pleno based schemes is better than our disparity compensation of super-rays. Also, the bitrate allocated to the segmentation and disparity is very large, especially at low bitrate (almost reaching $30$ percent for most datasets) and could be further reduced. 

Note that the decoder needs to compute the optimized basis functions for the non consistent super-rays, inducing some computational complexity. However, the optimization can be performed independently on each super-ray, in a parallel manner.

\section{Conclusion}
In this paper, we have addressed the problem of local geometry-aware graph transform design for light field energy compaction and compact representation. The transform support is based on super-rays constructed in a way that their shape remains coherent across the different views. We have first considered both non separable graph transforms. 

Despite the limited size of the transform support, the Laplacian matrix of such graph remains of high dimension and its diagonalization to compute the transform eigenvectors is computationally expensive. 

To solve this problem, we then considered a separable spatio-angular transform. We have shown that, when the shape of corresponding super-pixels in the different views undergoes small changes, the basis functions of the spatial transforms are not coherent, resulting in a decreased correlation between spatial transform coefficients. We hence proposed a novel transform optimization method that aims at preserving angular correlation even when the shapes of corresponding super-pixels (i.e. forming one super-ray) are not isometric. This procedure has been shown to increase energy compaction of the separable spatio-angular graph transforms and bring substantial rate-distortion performance gains compared to a non optimized case. 
The proposed optimized spatio-angular graph transforms can be applied on both color or residual signals and can be easily parallelized to reduce the complexity on the decoder side.  

\section*{Acknowledgment}
The authors would like to thank Elian Dib, Navid Mahmoudian Bidgoli and Pierre Allain from Inria for their help in extracting and running the JPEG Pleno VM 1.1 and HEVC CABAC encoder. Also, we would like to thank Effrosyni Simou and Eda Bayram from EPFL, who were of a great help with various discussions about the subject.

\section{Gradients of the objective function terms}
The gradients of the two terms in the optimization of equation \ref{eq:iter_opt_3} are provided below:
\begin{equation}
\begin{split}
& \nabla_B \Vert \mathbf{B}^\top \mathbf{\Lambda}_i \mathbf{B} - \mathbf{\Lambda}_i \Vert^2_F \\
&=\nabla_B tr \left((\mathbf{B}^\top\mathbf{\Lambda}_i\mathbf{B} - \mathbf{\Lambda}_i)^\top(\mathbf{B}^\top\mathbf{\Lambda}_i\mathbf{B} - \mathbf{\Lambda}_i)\right)\\
&= \nabla_B tr \left((\mathbf{B}^\top\mathbf{\Lambda}_i\mathbf{B} - \mathbf{\Lambda}_i^\top)(\mathbf{B}^\top\mathbf{\Lambda}_i\mathbf{B} - \mathbf{\Lambda}_i)\right)\\
&= \nabla_B tr (\mathbf{B}^\top\mathbf{\Lambda}_i\mathbf{B}\mathbf{B}^\top\mathbf{\Lambda}_i\mathbf{B} - \mathbf{B}^\top\mathbf{\Lambda}_i\mathbf{B}\mathbf{\Lambda}_i \\
& \qquad \qquad \qquad- \mathbf{\Lambda}_i^\top\mathbf{B}^\top\mathbf{\Lambda}_i\mathbf{B} + \mathbf{\Lambda}_i^\top\mathbf{\Lambda}_i)\\
&=4(\mathbf{\Lambda}_i \mathbf{B}\mathbf{B}^\top \mathbf{\Lambda}_i\mathbf{B} - \mathbf{\Lambda}_i \mathbf{B}\mathbf{\Lambda}_i)
\end{split}
\end{equation}

As for the coupling term, with a similar derivation as the first gradient and using the trace derivation properties in \cite{petersen2008matrix}, we get:
\begin{equation}
\begin{split}
&\nabla_B(\left\Vert(\mathbf{F}^\top \mathbf{U}_{s_0} - \mathbf{G}^\top \mathbf{U}_{s_i} \mathbf{B})\right\Vert^2_F) \\
&= 2 \mathbf{U}_{s_i}^\top\mathbf{G}(\mathbf{G}^\top\mathbf{U}_{s_i}\mathbf{B} - \mathbf{F}\mathbf{U}_{s_0})
\end{split}
\end{equation}

\bibliographystyle{IEEEtran}
\bibliography{ref}
%








\end{document}